\documentclass{article}

\usepackage{graphicx}%
\usepackage{multirow}%
\usepackage{amsmath,amssymb,amsfonts}%
\usepackage{amsthm}%
\usepackage{mathrsfs}%
\usepackage{makecell}
\usepackage[title]{appendix}%
\usepackage{xcolor}%
\usepackage{textcomp}%
\usepackage{manyfoot}%
\usepackage{booktabs}%
\usepackage{algorithm}%
\usepackage{algorithmicx}%
\usepackage{algpseudocode}%
\usepackage{listings}%
\usepackage{amsmath}
\usepackage{booktabs} 
\usepackage{multirow}
\usepackage[colorlinks=true,linkcolor=blue,urlcolor=blue]{hyperref}
\usepackage{graphicx}
\usepackage{subcaption}
\usepackage{booktabs}
\usepackage{multirow}

\usepackage[preprint]{neurips_2025}

\usepackage[utf8]{inputenc} 
\usepackage[T1]{fontenc}    
\usepackage{hyperref}       
\usepackage{url}            
\usepackage{booktabs}       
\usepackage{amsfonts}       
\usepackage{nicefrac}       
\usepackage{microtype}      
\usepackage{xcolor}         
\usepackage{graphicx}
\usepackage{subcaption}
\usepackage[export]{adjustbox} 
\usepackage{tabularx}

\title{Entropy-Driven Genetic Optimization for
Deep-Feature-Guided Low-Light Image Enhancement}

\author{%
  Nirjhor Datta$^{1,2}$ \\
  \And
  Afroza Akther$^{1,2}$ \\
  \And
  M. Sohel Rahman$^{1}$ \\
  \AND
  $^{1}$Department of Computer Science and Engineering, \\
  Bangladesh University of Engineering and Technology, \\
  West Palashi, Dhaka 1205, Bangladesh \\
  \And
  $^{2}$Department of Computer Science and Engineering, \\
  BRAC University, Dhaka 1205, Bangladesh \\
}
\begin{document}

\maketitle

\begin{abstract}
  Image enhancement methods often prioritize pixel level information, overlooking the semantic features. We propose a novel, unsupervised, fuzzy-inspired image enhancement framework guided by NSGA-II algorithm that optimizes image brightness, contrast, and gamma parameters to achieve a balance between visual quality and semantic fidelity. Central to our proposed method is the use of a pre-trained deep neural network as a feature extractor. 
 To find the best enhancement settings, we use a GPU-accelerated NSGA-II algorithm that balances multiple objectives, namely, increasing image entropy, improving perceptual similarity, and maintaining appropriate brightness. We further improve the results by applying a local search phase to fine-tune the top candidates from the genetic algorithm. Our approach operates entirely without paired training data making it broadly applicable across domains with limited or noisy labels. Quantitatively, our model achieves excellent performance with average BRISQUE and NIQE scores of 19.82 and 3.652, respectively, in all unpaired datasets. Qualitatively, enhanced images by our model exhibit significantly improved visibility in shadowed regions, natural balance of contrast and also preserve the richer fine detail without introducing noticable artifacts. This work opens new directions for unsupervised image enhancement where semantic consistency is critical. 
\end{abstract}

\section {Introduction}
Capturing high-quality images in low-light environments remains a long-standing challenge in computer vision. When illumination is insufficient, images often suffer from poor contrast, excessive noise, unnatural color shifts, and a loss of structural detail. These degradations not only reduce visual quality for human viewers but also hinder the performance of downstream tasks, such as object detection, recognition, and segmentation, particularly in critical applications like surveillance, autonomous driving, and underwater exploration. To tackle these challenges, a wide range of low-light image enhancement (LLIE) techniques have been proposed over the years \cite{source1,source10,source12,source13}.
\newline
Traditional LLIE approaches, such as histogram equalization and Retinex theory-based methods, attempt to correct lighting inconsistencies by assuming that an image can be decomposed into reflectance and illumination layers \cite{source17}. While these methods are conceptually appealing, they often struggle under real-world conditions with non-uniform lighting, frequently introducing artifacts, such as overexposure, halo effects, or unnatural colors. To address these shortcomings, learning-based techniques have gained prominence. For instance, LLNet employed deep autoencoders to learn enhancement mappings from data \cite{source15}, while the SID dataset introduced by Chen et al. \cite{source16} enabled supervised end-to-end learning directly from raw low-light images. More advanced methods, such as Retinex-Net \cite{source17} and KinD \cite{source18} combined deep learning with decomposition strategies, showing improved performance under controlled scenarios.
\newline
Despite these advances, several challenges remain. Traditional and early deep learning methods often face limitations, such as domain overfitting and poor generalization to varying lighting conditions. They may also struggle to preserve fine textures while effectively suppressing noise. Furthermore, many of these models require complex multi-stage training or rely on handcrafted design elements, which can limit their robustness and increase computational cost. To mitigate these issues, recent studies have explored new directions involving attention mechanisms, alternative color spaces, and Transformer-based architectures. For example, RetinexFormer \cite{source12} combines Retinex theory with a Transformer-based framework to better handle global feature modeling and reduce visual artifacts. MambaLLIE \cite{source10} introduces a memory-efficient state-space model that enhances both local detail and overall illumination. Meanwhile, the contrast-driven neural network (CDNN) proposed by Ryu et al. \cite{source1} tackles color distortion and over-enhancement through a multi-stage architecture specifically tuned for underwater imagery.
\\
Still, many open problems persist in this domain. Color inconsistency, over-enhancement, and the lack of perceptual alignment remain key obstacles. While metrics like PSNR and SSIM are widely used, they often fail to reflect human visual preferences \cite{source19}. More recent approaches, such as DCC-Net \cite{source14} have focused on improving color consistency, and PairLIE \cite{source7} have everaged paired instance learning to refine decomposition accuracy. However, achieving robust generalization and perceptual fidelity across diverse real-world scenes continues to be an unresolved challenge in low-light image enhancement research. 
\\
With the above backdrop, this paper makes the following key contributions.
\begin{itemize}
    \item We propose a novel image enhancement framework that leverages fuzzy logic-inspired adaptive operators (brightness, contrast, and gamma correction) to ensure gradual, perceptually driven improvements in image quality.
    \item We integrate high-level semantic awareness into the enhancement process by embedding a frozen deep feature extractor (e.g., VGG16).
    \item We design a multi-objective evaluation strategy that combines entropy maximization, perceptual similarity (feature loss), and luminance balance.
    \item We implement a GPU-accelerated NSGA-II to optimize enhancement parameters efficiently under multiple conflicting objectives.
    \item  Furthermore, we Integrate a local search mechanism to refine top-performing enhancement configurations for improved fine-tuning and convergence.
\end{itemize}

\section{Related Works}
Low-light image enhancement (LLIE) has witnessed significant progress in recent years, with the emergence of both traditional and modern techniques, such as deep learning-based methods. The goal remains consistent: to enhance visibility, preserve structural fidelity, and minimize visual artifacts in images captured under suboptimal lighting conditions. Previous works discuss the major developments across different paradigms, including Retinex-inspired methods, Transformer and state-space networks, Bayesian frameworks, fusion-based approaches, and color space transformations. In almost all works, evaluation metrics, such as PSNR, SSIM, LPIPS, NIQE, and BRISQUE are commonly used to benchmark performance across synthetic and real-world datasets. We will define these metrics in a forthcoming section.

\subsection{Deep Learning-Based Approaches}
Recent developments in low-light image enhancement have been driven largely by deep learning frameworks. These methods often focus on paired and unpaired training datasets, using metrics like PSNR, SSIM, LPIPS, NIQE, and BRISQUE to quantify performance.

Ryu et al.~\cite{source1} presented a novel approach for low-light image enhancement and color correction using a multi-stage deep learning method and a contrast-driven neural network. Their method addresses low illumination and contrast issues stemming from environmental factors like short shutter speed and atmospheric scattering. It is initially trained on the LOL dataset \cite{lol} and fine-tuned on underwater datasets \cite{source1} to mitigate problems like signal saturation, color distortion, and over-enhancement. The method demonstrated effectiveness in improving illuminance and contrast registering low BRISQUE scores.

Weng et al.~\cite{source10} proposed MambaLLIE, which utilized a global-local Mamba backbone that combines memory-efficient computation with impressive perceptual quality. This implicit Retinex-aware method employs a state space framework with feature stages, VSS blocks, and an SS2D mechanism.

Cai et al.~\cite{source12} introduced Retinexformer, a novel algorithm inspired by the Retinex theory and developed using Transformers. It featured an illumination-guided attention module and an ORF framework to handle noise and exposure artifacts effectively. The model achieved superior performance on thirteen datasets.

Zhang et al.~\cite{source14} proposed DCC-Net, a Deep Color Consistent Network, which used dual-camera input and Transformer architecture to jointly enhance illumination while maintaining natural color consistency. It performs competitively on multiple datasets.
Yan et al.~\cite{source13} developed CIDNet, which introduces a trainable HVI (Hue, Value, Intensity) color space that decouples brightness from color. The network, supported by Lighten Cross-Attention (LCA), demonstrated low computational cost and strong performance on various datasets.

Wu et al.~\cite{source3} presented Retinex-RNet, a method inspired by Retinex-Net that uses residual attention modules and is trained across multiple datasets for improved generalization. The model excels in exposure and local detail preservation.
Fu et al.~\cite{source7} introduced PairLIE, a framework trained using paired low-light samples. This method achieves accurate decomposition of illumination and reflectance components, validated through ablation studies on datasets like LOL and SICE.

\subsection{Fusion-Based and Multi-Exposure Techniques}
Fusion-based strategies are valued for their simplicity and adaptability. A fast LLIE technique, proposed in~\cite{source6}, uses multi-scale histogram fusion and bilateral filtering. Guo et al.~\cite{source6} employed metrics like NIQE and SSEQ to evaluate its efficiency. Karakata et al.~\cite{source11} proposed PAS-MEF, which fused multiple LDR inputs to simulate HDR output using PCA and saliency-based weight maps, mimicking human visual attention. On the other hand, Kadgirwar et al.~\cite{source4} introduced a Bayesian Multi-Exposure Image Fusion (MEF) method, incorporating probabilistic modeling in HDR reconstruction using Poisson-based frameworks.

\subsection{Traditional and Hybrid Techniques}
Traditional techniques remain relevant for scenarios requiring high computational efficiency. Alavi and Kargari~\cite{source2} enhanced each RGB channel independently, outperforming methods like HE and CLAHE on some datasets. 
Wu and Wu~\cite{source8} proposed an HSV-based method that applied power-law correction to the Value channel and merged enhancements through PCA, improving brightness and minimizing distortions.

A comparative summary of the key features, strengths, and evaluation benchmarks of LLIE methods is provided in Table~\ref{lit-summary} (Suppl. Section \ref{literature} in the Appendix) , highlighting both classical and modern approaches. 



\section{Methods}
\label{methods}

\begin{figure}[htbp]  
    \centering
    \includegraphics[width=\textwidth,height=5cm]{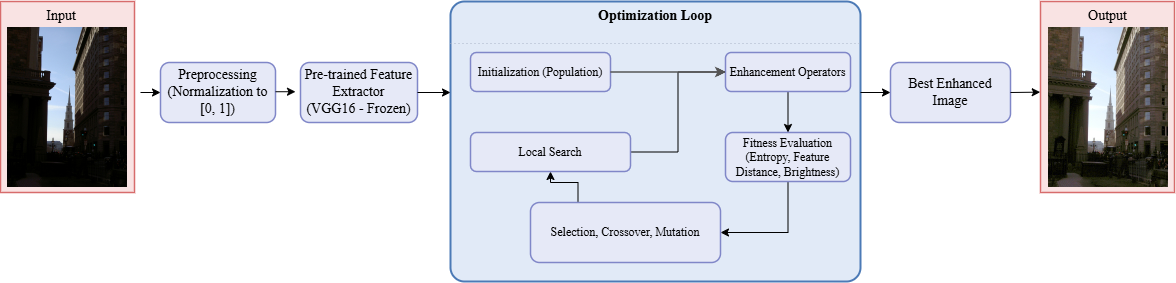}  
    \caption{Top-level diagram of the framework}
    \label{workflow}
\end{figure}

We formulate our image enhancement problem as a per-image bi-objective optimization problem over three continuous parameters, namely, brightness shift ($b$), contrast scaling ($c$), and gamma correction ($\gamma$) and solve it via a GPU accelerated hybrid NSGA-II algorithm \cite{NSGA} augmented with a local search. In this section, we detail the individual components of our proposed method along with their respective objectives. Figure \ref{workflow} presents an overview of our methods.

\subsection{Pretrained Feature Extractor}
We employ a pretrained VGG16 network~\cite{simonyan2014very} as a feature extractor. The network's convolutional layers are used to extract deep semantic features from the image. The classification head is removed, and the output of the last convolutional layer is pooled to a fixed size. Let \( I \) denote an input image, where \( I \in \mathbb{R}^{H \times W \times 3} \).
The feature extractor \( \mathcal{F} \) extracts features \( \mathbf{f} \in \mathbb{R}^C \) from input \( I \), where \( C \) is the dimension of the feature vector after adaptive pooling.
\subsection{Image Enhancement Operators}
The image enhancement is performed using fuzzy-inspired operations to adjust brightness, contrast, and gamma. Given an image \( I \), the enhancement is controlled by an individual represented by three parameters, namely, brightness, contrast, and gamma. These are denoted as \( \mathbf{e} = [b, c, g] \), where \( b \in [-10, 60] \), \( c \in [1, 2] \), and \( g \in [1,2] \). 
The brightness adjustment is performed by linearly shifting pixel values:
\[
I_{\text{bright}}(x, y)
= \operatorname{clamp}\!\Bigl(I(x, y) + \tfrac{b}{255},\,0,1\Bigr),
\]
where \(I(x,y)\) is the original (normalized) pixel value and \(b\in[-30,30]\) is the brightness parameter. 
Contrast scaling is then applied to the brightness-shifted image:
\[
I_{\text{contrast}}(x, y)
= \operatorname{clamp}\!\Bigl(c \,\cdot\, I_{\text{bright}}(x, y),\,0,1\Bigr),
\]
where \(c\in[0.8,1.2]\) is the contrast factor. 
Finally, gamma correction controls the overall intensity curve via:
\[
I_{\gamma}(x, y)
= \operatorname{clamp}\!\bigl(I_{\text{contrast}}(x, y)^{\,1/\gamma},\,0,1\bigr),
\]
where \(\gamma\in[0.8,1.2]\) is the gamma parameter.

\subsection{Fitness Evaluation: Multi-Objective Function}
The fitness function is designed to optimize three objectives:
\begin{description}
    \item[Enhancement Quality.] This is measured using Shannon entropy of the grayscale image to quantify the level of detail in the enhanced image.
    \item[Semantic Feature Preservation.] This is measured as the Euclidean distance between the feature vector of the enhanced image and the original image.
    \item[Brightness Constraint Penalty.] A penalty is applied if the enhanced image's brightness falls below a threshold \( \theta_{\text{bright}} \).
\end{description}

Given an individual \( \mathbf{e} = [b, c, g] \) and the original image \( I_0 \), the fitness is computed as follows:
\[
\text{Fitness}(\mathbf{e}) =
\left( 
\begin{array}{c}
-\text{Entropy}(I_{\text{enhanced}}), \\
\text{FeatureDistance}(I_0, I_{\text{enhanced}}) 
+ \lambda \cdot \text{BrightnessPenalty}(I_{\text{enhanced}})
\end{array}
\right)
\]
Here, \( I_{\text{enhanced}} \) represents the enhanced image generated using parameters \( \mathbf{e} \), while \( \lambda \) is a scaling factor applied to the brightness penalty. The entropy of an image \( I_{\text{enhanced}} \) is computed as \( \text{Entropy}(I) = -\sum_{i=0}^{255} p(i) \log_2 p(i) \), where \( p(i) \) denotes the normalized histogram of the grayscale version of the image. To assess perceptual consistency, the feature distance is calculated using the Euclidean distance between the feature vectors of the original and enhanced images, given by \( \text{FeatureDistance}(I_0, I_{\text{enhanced}}) = \|\mathcal{F}(I_0) - \mathcal{F}(I_{\text{enhanced}})\|_2 \). Additionally, a brightness penalty is imposed to prevent overly dark outputs, defined as \( \text{BrightnessPenalty}(I) = \max(0, \theta_{\text{bright}} - \text{MeanBrightness}(I)) \cdot \eta \), where \( \eta \) is a penalty scaling factor and \( \text{MeanBrightness}(I) \) denotes the mean pixel intensity of image \( I \).

\subsection{Adaptive Parameter Control}
To control the mutation rate adaptively during the evolutionary process, we compute the standard deviation of the brightness parameter from the current population as follows.
\[
\sigma_{\text{brightness}} = \text{std}(\{e_1[0], e_2[0], \dots, e_n[0]\})
\]
If \( \sigma_{\text{brightness}} \) is low, indicating low diversity in the population, the mutation rate is increased to promote exploration. The mutation rate is also adaptive as follows.
\[
r_{\text{mutate}} =
\begin{cases}
\min(0.5, 2 \cdot r_{\text{base}}), & \text{if } \sigma_{\text{brightness}} < 5 \\
r_{\text{base}}, & \text{otherwise}
\end{cases}
\]
Here \( r_{\text{base}} \) is the base mutation rate.

\subsection{Memetic Local Search}
In order to improve the quality of the individuals in the population, a local search is applied to the top 10\% of the population according to fitness. Given an individual \( \mathbf{e} \), the neighbor is generated by making small perturbations to each parameter as follows.
\[
\mathbf{e}_{\text{neighbor}} = \mathbf{e} + \delta, \quad \delta \sim \mathcal{N}(0, \sigma^2),
\]
where \( \sigma \) is a small value. The neighbor is accepted only if its fitness is improved, i.e., if $\text{Fitness}(\mathbf{e}_{\text{neighbor}}) < \text{Fitness}(\mathbf{e})$ 

\subsection{NSGA-II Based Multi-Objective Optimization}
Our approach employs the NSGA-II algorithm \cite{NSGA} (Suppl. Section \ref{nsga} in the Appendix) to optimize image enhancement parameters by simultaneously maximizing entropy and minimizing deep feature distance with respect to the original image (Algorithm \ref{algo1}). 
Although NSGA-II evolves a Pareto front of non-dominated solutions over \((f_1, f_2)\), we ultimately require a single parameter set for each image. Therefore, after the final generation, we extract the rank-0 front and select the individual with the highest entropy (and, in the case of a tie, the one with the lowest sum of feature distance and brightness penalty). This corresponds to \texttt{selBest(pop, 1)} in the DEAP framework applied to a \texttt{FitnessMulti} object, which first filters by Pareto rank and then by crowding distance. Thus, in effect, we select one representative solution from the front by choosing the individual with the highest entropy and the lowest feature distortion to ensure the balance of visual quality and semantic similarity. 

\begin{algorithm}[H]
\caption{NSGA-II Based Multi-Objective Deep Feature-Guided Image Enhancement}
\label{algo1}
\begin{algorithmic}[1]
\State Initialize $P_0$ with random $e=[b,c,\gamma]$
\For{$g=1$ \dots $G$}
  \State Evaluate $(f_1,f_2)$ for each $e\in P_g$
  \State $P_g \gets$ NSGA-II\_select($P_g$)        \hfill\(\triangleright\) nondominated + crowding
  \State Generate offspring via crossover/mutation
  \State Evaluate offspring, form $P_{g+1}$ via NSGA-II
  \State Apply local search on top 10\% of $P_{g+1}$ 
 selected based on fitness
\EndFor
\State \(\mathcal{P}^*\leftarrow\) Pareto front of $P_G$
\State Sort \(\mathcal{P}^*\) by \(\bigl(\,-f_1,\;f_2\bigr)\)
\State \Return the first element of \(\mathcal{P}^*\)
\end{algorithmic}
\end{algorithm}

\begin{table}[ht]
\centering
\caption{Hyperparameter Settings}
\label{tab:hyper}
\begin{adjustbox}{width=0.7\textwidth}
\begin{tabular}{lcccccc}
\toprule
\textbf{$N$} & \textbf{$G$} & \textbf{$p_c$} & \textbf{$p_m$} & \textbf{$I_{\mathrm{LS}}$} & \textbf{$\lambda$} & \textbf{$[m \pm \delta]$} \\
\midrule
50 & 5 & 0.85 & 0.3\,{$\rightarrow$}\,0.2 & 8 & 30 & [0.35,\,0.7] \\
 \\
\bottomrule
\end{tabular}
\end{adjustbox}
\end{table}

\subsection{Hyperparameter Selection}
The hyperparameters of our proposed method are presented at Table \ref{tab:hyper}. These values were carefully selected through first-principles reasoning and domain-specific considerations. \textbf{Population size} (N=50) balances exploration-exploitation trade-offs while maintaining computational feasibility. \textbf{Five generations} were chosen to prevent overfitting while allowing meaningful convergence, observing diminishing returns beyond this threshold. A high crossover rate (0.85) promotes solution-space exploration, complemented by an adaptive mutation rate (0.3$\rightarrow$0.2) that transitions from broad exploration to localized refinement. \textbf{Eight local-search steps} enable sufficient neighborhood exploitation without excessive computation. The \textbf{brightness penalty} ($\lambda$=30) was scaled to match feature-space distances, while bounds [0.35,0.65] prevent information loss in extreme luminance ranges, aligning with human visual perception thresholds. These choices collectively optimize solution quality under practical computational constraints.
These parameters were chosen as conservative first estimates, informed by established heuristics in image processing and foundational principles of evolutionary computation. 
\subsection{Code, Environment and Availability}
\label{code}
All
the models are trained on a A6000 GPU 48 GB, RAM: 64GB DDR5 SSD: 1
TB PSU: 1000W PC. The code is available \href{https://github.com/NIRJHOR-DATTA/Entropy-Driven-Genetic-Optimization-for-Deep-Feature-Guided-Low-Light-Image-Enhancement}{\texttt{here}}. The environment was built using Python 3.10. Additionally, we have used \texttt{NumPy} 2.1.0, \texttt{OpenCV} 4.11.0, \texttt{matplotlib} 3.10.1, \texttt{PyTorch} 2.5.1, \texttt{torchvision} 0.20.1, \texttt{scikit-image} 0.25.2 and \texttt{deap} 1.4 (for Genetic Algorithm implementation).
\section{Results}
\subsection{Datasets}
\label{datasetsused}
We have performed extensive experimental evaluation from two different angles. Firstly, we evaluate how our approach works for the unpaired images. Since our proposed method do not require ground truth images, we have experimented with five commonly used unpaired image LLIE datasets for evaluation, namely,  DICM \cite{DICM} , LIME\cite{LIME} , MEF \cite{MEF} , NPE \cite{NPE}, and VV \cite{VV} (Table \ref{tab:datasets}).
\begin{table}[ht]
\centering
\small
\caption{Summary of Low-Light Image Datasets Used}
\label{tab:datasets}
\begin{tabular}{|l|c|p{6.3cm}|c|}
\hline
\textbf{Dataset} & \textbf{No. of Images} & \textbf{Image Type} & \textbf{Format} \\
\hline
\textbf{DICM} & 69 & Real-world low-light images (night streets, cars, underwater flowers, shadows, dark rooms) & JPG \\
\hline
\textbf{LIME} & 10 & Natural scenes captured under low lighting & BMP \\
\hline
\textbf{MEF} & 17 & Indoor and outdoor natural scenes in low light & PNG (High-res) \\
\hline
\textbf{NPE} & 8 & Outdoor scenes under poor lighting conditions & JPG \\
\hline
\textbf{VV} & 24 & Daytime outdoor scenes in poorly lit environments (urban streets, vehicles, buildings) & JPG \\
\hline
\end{tabular}
\end{table}
Secondly, we have compared the performance of CIDNet \cite{source13} \cite{fediory2025hvi_cidnet} (with their published generalized weights) against our model on 100 randomly selected images of MIT-5K dataset that contains 5000 raw images and 5 retouched versions thereof.

\subsection{Evaluation Metrics}
Evaluation is challenging due to the absence of ground truth in real-world data. Hou et al.~\cite{source9} used PSNR, SSIM, and LPIPS on paired datasets like LOLv1 and LOLv2, while using NIQE for unpaired datasets like DICM and LIME. Puzovic et al.~\cite{source5} emphasized objective metrics for robust evaluation. Recent models like MambaLLIE~\cite{source10} argue against the sufficiency of PSNR and SSIM, instead leverage perceptual prompts and vision-language models like CLIP-IQA for human-aligned quality assessments. Based on above, for the unpaired datasets, we have evaluated using BRISQUE \cite{BRISQUE} and NIQE \cite{NIQE}, where the former evaluates the perceptual quality of the natural images based on spatial domain statistics and the latter uses statistical features of natural scenes without relying on human level training data.
On the other hand, for the second task, 
we have used Peak Signal-to-Noise Ratio (PSNR), Structural Similarity (SSIM) and Entropy.
The Peak Signal-to-Noise Ratio (PSNR) is calculated as \( \text{PSNR} = 10 \log_{10} \left( \frac{\text{MAX}^2_I}{\text{MSE}} \right) \), where \( \text{MAX}_I \) is the maximum possible pixel value (e.g., 255 for 8-bit images) and MSE is the Mean Squared Error. The Structural Similarity Index (SSIM) is given by \( \text{SSIM}(x, y) = \frac{(2\mu_x\mu_y + C_1)(2\sigma_{xy} + C_2)}{(\mu_x^2 + \mu_y^2 + C_1)(\sigma_x^2 + \sigma_y^2 + C_2)} \), where \( \mu_x \) and \( \mu_y \) are the mean values of images \( x \) and \( y \), \( \sigma_x^2 \) and \( \sigma_y^2 \) are the variances, \( \sigma_{xy} \) is the covariance, and \( C_1, C_2 \) are small constants to stabilize the equation.
\subsection{Results on unpaired datasets}
\begin{figure}[h!]  
    \centering
    \includegraphics[width=10cm,height=7cm]{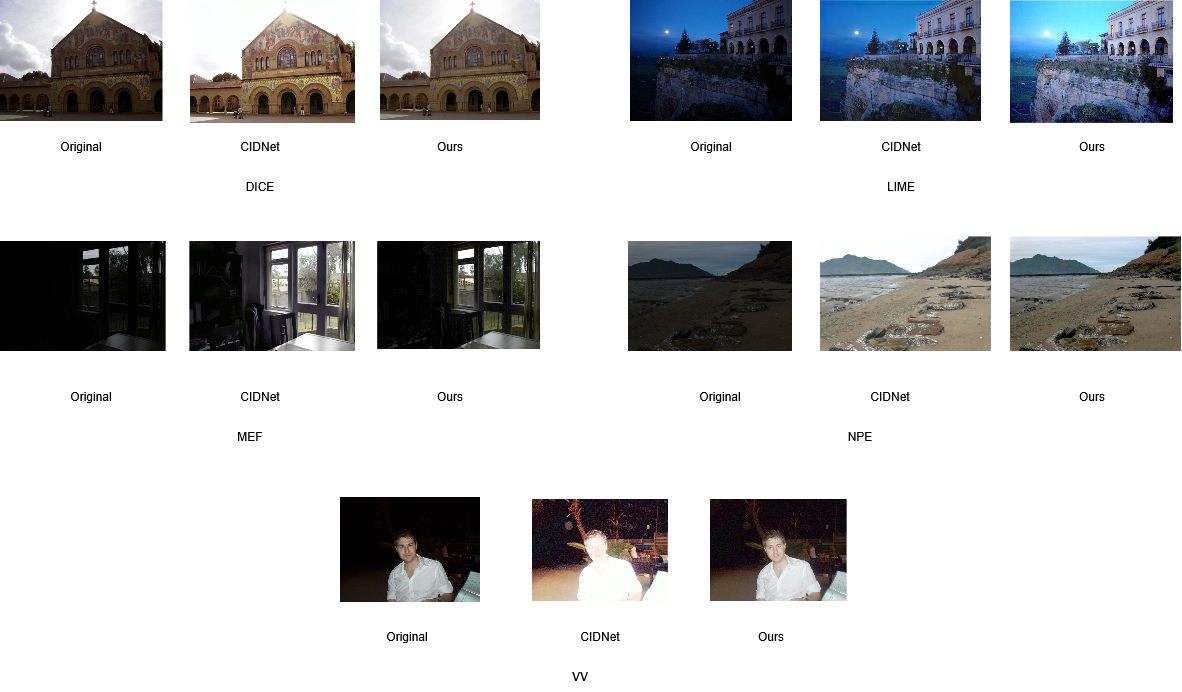}  
    \caption{Visual comparison of enhancement results by our method versus CIDNet on a random sample from each dataset}
    \label{results}
\end{figure}

We evaluate the performance of our model on 5 unpaired datasets using BRISQUE and NIQE metric. Lower value in both metrics mean higher perceptual qualty. 
The results are presented in Table \ref{tab:main_ablation}, where we can observe that our model outperforms other models in the DICM dataset and VV dataset. While our approach does not outperform other models in the BRISQUE metric in NPE dataset, it outperfoms other models in NIQE metric which reflects that our approach's recovered perceptual results are closer to realistic appearances than others. Figure \ref{results} presents a visual comparison between the produced images (one random sample from each dataset) by our model and that by CIDNet \cite{source13}. Further visual comparison is available in the Appendix (Suplementary Section \ref{visualizeunpaired}, Fig. \ref{fig:examples_dicm} -\ref{fig:examples_vv}).

\begin{table}[ht]
\centering
\small
\caption{BRISQUE and NIQE scores ($\downarrow$ lower is better) on five unpaired datasets, with ablated variants. Lowest values in each column are \textbf{bolded}.}
\label{tab:main_ablation}
\resizebox{\textwidth}{!}{%
\begin{tabular}{l*{10}{c}}
\toprule
\multirow{2}{*}{\textbf{Method}} 
  & \multicolumn{2}{c}{\textbf{DICM}} 
  & \multicolumn{2}{c}{\textbf{LIME}} 
  & \multicolumn{2}{c}{\textbf{MEF}} 
  & \multicolumn{2}{c}{\textbf{NPE}} 
  & \multicolumn{2}{c}{\textbf{VV}} \\
\cmidrule(lr){2-3}\cmidrule(lr){4-5}\cmidrule(lr){6-7}\cmidrule(lr){8-9}\cmidrule(lr){10-11}
  & BRISQUE & NIQE
  & BRISQUE & NIQE
  & BRISQUE & NIQE
  & BRISQUE & NIQE
  & BRISQUE & NIQE \\
\midrule
KinD \cite{source18}            & 48.72 & 5.15 & 39.91 & 5.03 & 49.94 & 5.47 & 36.85 & 4.98 & 50.56 & 4.30 \\
ZeroDCE \cite{ZeroDCE}          & 27.56 & 4.58 & 20.44 & 5.82 & 17.32 & 4.93 & 20.72 & 4.53 & 34.66 & 4.81 \\
RUAS \cite{RUAS}            & 38.75 & 5.21 & 27.59 & 4.26 & 23.68 & 3.83 & 47.85 & 5.53 & 38.37 & 4.29 \\
LLFlow \cite{LLFlow}          & 26.36 & 4.06 & 27.06 & 4.59 & 30.27 & 4.70 & 28.86 & 4.67 & 31.67 & 4.04 \\
SNR-Aware \cite{SNR-Aware}       & 37.35 & 4.71 & 39.22 & 5.74 & 31.28 & 4.18 & 26.65 & 4.32 & 78.72 & 9.87 \\
PairLIE \cite{source7}          & 33.31 & 4.03 & 25.23 & 4.58 & 27.53 & 4.06 & 28.27 & 4.18 & 39.13 & 3.57 \\
CIDNet \cite{source13}          & 21.47 & 3.79 & \textbf{16.25} & \textbf{4.13} & 13.77 & 3.56 & 18.92 & 3.74 & 30.63 & 3.21 \\
Ours (Full Model)    & 20.79 & \textbf{3.73} & 21.76 & 4.43 & \textbf{12.79} & 3.40 & 17.85 & 3.78 & \textbf{25.91} & \textbf{2.92} \\
\midrule
\multicolumn{11}{l}{\emph{Ablated Versions}} \\
NSGA-II only         & 20.40 & \textbf{3.70} & 20.88 & 4.25 & 12.84 & \textbf{3.38}&  \textbf{17.67} & 3.78 &  26.94 & 3.05 \\
No local search      &  20.41  & 3.73 & 21.67 & 4.37  &  12.93 & 3.42 & 17.93 & \textbf{3.72} &  26.22 & 3.00 \\
No adaptive mutation &  \textbf{20.39} & 3.73 &  22.07 & 4.38 &  13.04& 3.40 &  18.15 & 3.74 &  26.37 & 2.99 \\
\bottomrule
\end{tabular}%
}
\end{table}

\subsection{Results on MIT-5K datasets}
\begin{figure}[h!]
    \centering
    \includegraphics[width=\linewidth]{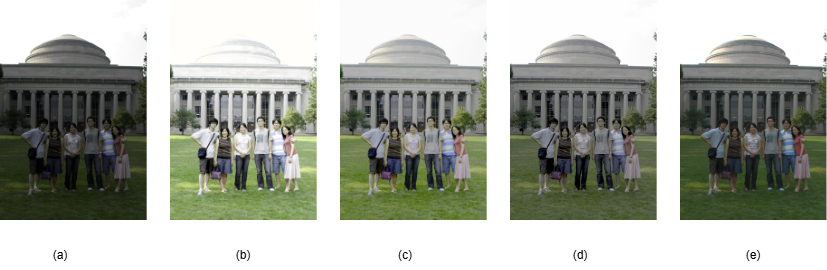}  
    \caption{Visual comparison of all the approaches-(a)-retouched version ,(b)-CIDNet lolv1 weight , (c)-CIDNet generalization weight, (d)-ours, (e)- actual image}
    \label{paired}
\end{figure}

As can be observed from the results reported in Table \ref{paired}, our model achieves the highest PSNR in all five retouched versions of this dataset, indicating superior reconstruction capability. Structural similarity is also consistently better in our method, which means that the enhanced image has maintained structural similarity to the ground truth image. Our method demonstrates higher entropy values across all five retouched versions, suggesting it produces richer and more natural-looking images with more information content. 
Figure \ref{paired} presents a visual comparison between a visual comparison of enhancement results between our apporach and different CIDNet \cite{source13} weight configurations.

\begin{table}[ht]
\centering
\caption{PSNR, SSIM, and Entropy on MIT-5K Subsets}
\label{tab:mit5k_metrics}
\small  
\begin{adjustbox}{width=0.6\textwidth}
\begin{tabular}{clccc}
\toprule
\textbf{Subset} & \textbf{Approach} & \textbf{PSNR} & \textbf{SSIM} & \textbf{Entropy} \\
\midrule
\multirow{3}{*}{a} & CIDNet-lolv1             & 9.7833  & 0.4502 & 7.0637 \\
                   & CIDNet-generalization    & 14.3492 & 0.5179 & 7.3831 \\
                   & Ours                     & 15.8024 & 0.5413 & 7.6369 \\
\midrule
\multirow{3}{*}{b} & CIDNet-lolv1             & 8.8100  & 0.4483 & 6.9222 \\
                   & CIDNet-generalization    & 13.7503 & 0.5237 & 7.2718 \\
                   & Ours                     & 15.9918 & 0.5627 & 7.6397 \\
\midrule
\multirow{3}{*}{c} & CIDNet-lolv1             & 9.7833  & 0.4502 & 7.0637 \\
                   & CIDNet-generalization    & 8.8172  & 0.4372 & 6.9389 \\
                   & Ours                     & 15.0147 & 0.5353 & 7.6791 \\
\midrule
\multirow{3}{*}{d} & CIDNet-lolv1             & 8.2457  & 0.4502 & 7.0637 \\
                   & CIDNet-generalization    & 14.0834 & 0.4364 & 6.6912 \\
                   & Ours                     & 15.5798 & 0.5516 & 7.6060 \\
\midrule
\multirow{3}{*}{e} & CIDNet-lolv1             & 8.1494  & 0.4251 & 6.7677 \\
                   & CIDNet-generalization    & 13.3932 & 0.5057 & 7.3300 \\
                   & Ours                     & 14.7874 & 0.5283 & 7.6903 \\
\bottomrule
\end{tabular}
\end{adjustbox}
\end{table}

\subsection{Ablation Study}
Table~\ref{tab:main_ablation} reports BRISQUE and NIQE scores for each ablation variant for the five unpaired datasets. 
Interestingly, the NSGA-II only variant achieves the lowest NIQE  (3.7041), while the no adaptive mutation variant achieves the lowest BRISQUE (20.3859) in the DICM dataset. The full model remains competitive, but does not strictly dominate on both metrics. NSGA-II only variant achieves the best scores on both BRISQUE (20.8784) and NIQE (4.2541) on the LIME dataset as well, indicating that for this simple dataset NSGA-II actually suffices. Interestingly, NSGA-II only variant still is beaten by CIDNet in terms of both BRISQUE and NIQE, but only slightly in case of the former.
The NSGA-II only variant attains the lowest NIQUE (3.3776) and the full model achieves the lowest BRISQUE (12.7930) in the MEF dataset.
Removing local search produces the best NIQUE (3.7209) even beating the CIDNet, while the NSGA-II only variant gives the best BRISQUE (17.6708) in the NPE dataset, again showing that individual enhancements can sometimes arise without memetic refinement. The full model achieves the best performance on both NIQUE (2.9245) and BRISQUE (25.9100) on VV dataset, demonstrating that on more diverse and challenging scenes, the combination of adaptive mutation and local search yields a clear advantage. Overall, although the NSGA-II only variant often matches or slightly outperforms individual components on simpler datasets, the full model is the most robust especially on the heterogeneous VV dataset validating the utility of both adaptive mutation and memetic local search for complex low-light enhancement tasks.
\section{Discussions}
\label{discussion}
We have achieved superior performance in terms of BRISQUE (NIQE) on four (three) out of five unpaired datasets (Table \ref{tab:main_ablation}). In terms of BRISQUE (NIQE) the supervised network, CIDNet outperforms our model in one, namely, LIME  (two, namely, LIME and NPE) dataset(s). Interestingly, in NPE, an ablated version of our model outperforms CIDNet in terms of NIQE. 
NSGA-II in our approach optimizes global brightness, contrast, and gamma in each image independently. Our approach maintains a balanced contrast and preserves fine details across diverse lightning conditions. As shown in Figure \ref{results},  CIDNet \cite{source13} tends to overexpose certain regions, leading to loss of structural information, whereas our method effectively enhances brightness without sacrificing local contrast or introducing visual artifacts. On scenes requiring highly localized corrections (e.g., very small bright spots or complex shadows), the three-parameter model may be too large, leading to under or over correction in some regions. This might be one of the reasons why our model could not beat a supervised machine learning model on those two datasets. \\
On the paired dataset (Table~\ref{paired}), our model remains competitive with supervised approaches despite never seeing any training data because the fitness function directly optimizes perceptual entropy and deep feature consistency. As seen in Figure \ref{paired}, our method produces visually balanced enhancements with better preservation of natural contrast, avoiding the over enhancement artifacts observed in other CIDNet \cite{source13} variants which is consistent with our superior PSNR and SSIM scores.
Supervised methods benefit from end-to-end fine-tuning on the exact imaging conditions, which explains their slight edge in certain pixel-wise metrics (PSNR/SSIM); still our model was able to defeat these models in terms of PSNR and SSIM which speaks volumes about the effectiveness thereof.

This brings us to the most significant advantage of our approach, which is the zero training requirement.  Unlike supervised or GAN-based enhancers, we do not need large collections of paired or unpaired examples, nor expensive GPU hours for network training or fine-tuning. 
Another important factor is that users can inspect or constrain the image enhancement operator's ranges, insert additional penalty terms, or modify the fitness criteria, which is difficult in black-box neural networks. By optimizing each image independently, our method avoids cumulative degradation when test data is different from any fixed training distribution.

Along with these advantages, our proposed approach comes with certain limitations too. Due to its iterative nature, the proposed method might not be suited for real time enhancements without significant optimization or hardware acceleration. Our optimization is entirely image statistics (entropy, feature distance, brightness) driven. As a result it lacks semantic understanding which means it does not account for high level features like object identity, facial quality or scene context. Unlike deep learning models it does not learn from data distributions over time, as a result, its adaptability to unseen domains may be limited.   
\section{Conclusion}
In this paper, we proposed a novel fuzzy-inspired image enhancement framework that leverages NSGA-II and pre-trained deep features to produce perceptually superior images while preserving semantic content. Unlike conventional enhancement techniques that often rely on fixed or heuristic-based parameter tuning, our method adaptively searches for optimal enhancement parameters (brightness, contrast, and gamma) using a multi-objective optimization process. By incorporating entropy, deep feature similarity, and brightness constraints into the fitness function, our approach ensures both information preservation and visual clarity. Experimental results demonstrate that our model achieves state-of-the-art performance on 3 out of 5 unpaired datasets, while also exhibiting competitive effectiveness on paired datasets outperforming a supervised enhancement method in different settings despite operating in an unsupervised manner.
\bibliography{reference}

\begin{thebibliography}{10}

\bibitem{source2}
Meysam Alavi and Mehrdad Kargari.
\newblock A new contrast enhancement method for color dark and low-light images.
\newblock In {\em 2022 9th Iranian Joint Congress on Fuzzy and Intelligent Systems (CFIS)}, page 1–7. IEEE, March 2022.

\bibitem{source12}
Yuanhao Cai, Hao Bian, Jing Lin, Haoqian Wang, Radu Timofte, and Yulun Zhang.
\newblock Retinexformer: One-stage retinex-based transformer for low-light image enhancement.
\newblock In {\em Proceedings of the IEEE/CVF International Conference on Computer Vision (ICCV)}, pages 12504--12513, October 2023.

\bibitem{source16}
Chen Chen, Qifeng Chen, Jia Xu, and Vladlen Koltun.
\newblock Learning to see in the dark.
\newblock In {\em Proceedings of the IEEE Conference on Computer Vision and Pattern Recognition (CVPR)}, June 2018.

\bibitem{NSGA}
K.~Deb, A.~Pratap, S.~Agarwal, and T.~Meyarivan.
\newblock A fast and elitist multiobjective genetic algorithm: Nsga-ii.
\newblock {\em IEEE Transactions on Evolutionary Computation}, 6(2):182--197, 2002.

\bibitem{fediory2025hvi_cidnet}
{Fediory}.
\newblock {HVI-CIDNet} · hugging face.
\newblock \url{https://huggingface.co/Fediory/HVI-CIDNet}, 2025.
\newblock Accessed: 2025-05-12.

\bibitem{source7}
Zhenqi Fu, Yan Yang, Xiaotong Tu, Yue Huang, Xinghao Ding, and Kai-Kuang Ma.
\newblock Learning a simple low-light image enhancer from paired low-light instances.
\newblock In {\em Proceedings of the IEEE/CVF Conference on Computer Vision and Pattern Recognition (CVPR)}, pages 22252--22261, June 2023.

\bibitem{ZeroDCE}
Chunle Guo, Chongyi Li, Jichang Guo, Chen~Change Loy, Junhui Hou, Sam Kwong, and Runmin Cong.
\newblock Zero-reference deep curve estimation for low-light image enhancement.
\newblock In {\em Proceedings of the IEEE/CVF Conference on Computer Vision and Pattern Recognition (CVPR)}, June 2020.

\bibitem{source6}
Junjie Guo, Zhengping Li, Chao Xu, and Bo~Feng.
\newblock Fast low-light image enhancement algorithm based on fusion.
\newblock In {\em 2021 IEEE 6th International Conference on Signal and Image Processing (ICSIP)}, page 251–256. IEEE, October 2021.

\bibitem{LIME}
Xiaojie Guo, Yu~Li, and Haibin Ling.
\newblock Lime: Low-light image enhancement via illumination map estimation.
\newblock {\em IEEE Transactions on Image Processing}, 26(2):982--993, 2017.

\bibitem{source9}
Jinhui HOU, Zhiyu Zhu, Junhui Hou, Hui LIU, Huanqiang Zeng, and Hui Yuan.
\newblock Global structure-aware diffusion process for low-light image enhancement.
\newblock In A.~Oh, T.~Naumann, A.~Globerson, K.~Saenko, M.~Hardt, and S.~Levine, editors, {\em Advances in Neural Information Processing Systems}, volume~36, pages 79734--79747. Curran Associates, Inc., 2023.

\bibitem{source11}
Diclehan Karakaya, Oguzhan Ulucan, and Mehmet Turkan.
\newblock Pas-mef: Multi-exposure image fusion based on principal component analysis, adaptive well-exposedness and saliency map.
\newblock In {\em ICASSP 2022 - 2022 IEEE International Conference on Acoustics, Speech and Signal Processing (ICASSP)}, pages 2345--2349, 2022.

\bibitem{DICM}
Chulwoo Lee, Chul Lee, and Chang-Su Kim.
\newblock Contrast enhancement based on layered difference representation of 2d histograms.
\newblock {\em IEEE Transactions on Image Processing}, 22(12):5372--5384, 2013.

\bibitem{RUAS}
Risheng Liu, Long Ma, Jiaao Zhang, Xin Fan, and Zhongxuan Luo.
\newblock Retinex-inspired unrolling with cooperative prior architecture search for low-light image enhancement.
\newblock In {\em Proceedings of the IEEE/CVF Conference on Computer Vision and Pattern Recognition (CVPR)}, pages 10561--10570, June 2021.

\bibitem{source15}
Kin~Gwn Lore, Adedotun Akintayo, and Soumik Sarkar.
\newblock Llnet: A deep autoencoder approach to natural low-light image enhancement.
\newblock {\em Pattern Recognition}, 61:650--662, 2017.

\bibitem{MEF}
Kede Ma, Kai Zeng, and Zhou Wang.
\newblock Perceptual quality assessment for multi-exposure image fusion.
\newblock {\em IEEE Transactions on Image Processing}, 24(11):3345--3356, 2015.

\bibitem{BRISQUE}
Anish Mittal, Anush~Krishna Moorthy, and Alan~Conrad Bovik.
\newblock No-reference image quality assessment in the spatial domain.
\newblock {\em IEEE Transactions on Image Processing}, 21(12):4695--4708, 2012.

\bibitem{NIQE}
Anish Mittal, Rajiv Soundararajan, and Alan~C. Bovik.
\newblock Making a “completely blind” image quality analyzer.
\newblock {\em IEEE Signal Processing Letters}, 20(3):209--212, 2013.

\bibitem{source5}
Snezana Puzovic, Ranko Petrovic, Milos Pavlovic, and Srdan Stankovic.
\newblock Enhancement algorithms for low-light and low-contrast images.
\newblock In {\em 2020 19th International Symposium INFOTEH-JAHORINA (INFOTEH)}, page 1–6. IEEE, March 2020.

\bibitem{source4}
Qiu Rui, Jingyu Wang, Pierre Thibault, Xin Zhang, Hau-Tieng Wu, Jean-Baptiste Sibarita, and Guo Zhang.
\newblock Bayesian multi-exposure image fusion for robust high dynamic range ptychography.
\newblock {\em Advanced Photonics Nexus}, 2(6):066002, 2023.

\bibitem{source1}
Jaeseok Ryu, Heunseung Lim, Hyeongseok Oh, Jeonghak Oh, and Joonki Paik.
\newblock Low-light image enhancement and color correction using a contrast-driven neural network.
\newblock In {\em 2025 IEEE International Conference on Consumer Electronics (ICCE)}, page 1–3. IEEE, January 2025.

\bibitem{simonyan2014very}
Karen Simonyan.
\newblock Very deep convolutional networks for large-scale image recognition.
\newblock {\em arXiv preprint arXiv:1409.1556}, 2014.

\bibitem{VV}
Vassilios Vonikakis, Rigas Kouskouridas, and Antonios Gasteratos.
\newblock On the evaluation of illumination compensation algorithms.
\newblock {\em Multimedia Tools and Applications}, 77(8):9211–9231, May 2017.

\bibitem{NPE}
Shuhang Wang, Jin Zheng, Hai-Miao Hu, and Bo~Li.
\newblock Naturalness preserved enhancement algorithm for non-uniform illumination images.
\newblock {\em IEEE Transactions on Image Processing}, 22(9):3538--3548, 2013.

\bibitem{LLFlow}
Yufei Wang, Renjie Wan, Wenhan Yang, Haoliang Li, Lap-Pui Chau, and Alex Kot.
\newblock Low-light image enhancement with normalizing flow.
\newblock {\em Proceedings of the AAAI Conference on Artificial Intelligence}, 36(3):2604–2612, June 2022.

\bibitem{source17}
Chen Wei, Wenjing Wang, Wenhan Yang, and Jiaying Liu.
\newblock Deep retinex decomposition for low-light enhancement, 2018.

\bibitem{lol}
Chen Wei, Wenjing Wang, Wenhan Yang, and Jiaying Liu.
\newblock Deep retinex decomposition for low-light enhancement, 2018.

\bibitem{source10}
Jiangwei Weng, Zhiqiang Yan, Ying Tai, Jianjun Qian, Jian Yang, and Jun Li.
\newblock Mamballie: Implicit retinex-aware low light enhancement with global-then-local state space, 2024.

\bibitem{source8}
Na~Wu and Yapeng Wu.
\newblock Low-illumination color image enhancement algorithm based on hsv space.
\newblock In {\em 2025 IEEE 8th Information Technology and Mechatronics Engineering Conference (ITOEC)}, page 1523–1528. IEEE, March 2025.

\bibitem{source3}
Qi~Wu, Maoling Qin, Jingqi Song, and Li~Liu.
\newblock An improved method of low light image enhancement based on retinex.
\newblock In {\em 2021 6th International Conference on Image, Vision and Computing (ICIVC)}, page 233–241. IEEE, July 2021.

\bibitem{SNR-Aware}
Xiaogang Xu, Ruixing Wang, Chi-Wing Fu, and Jiaya Jia.
\newblock Snr-aware low-light image enhancement.
\newblock In {\em Proceedings of the IEEE/CVF Conference on Computer Vision and Pattern Recognition (CVPR)}, pages 17714--17724, June 2022.

\bibitem{source13}
Qingsen Yan, Yixu Feng, Cheng Zhang, Pei Wang, Peng Wu, Wei Dong, Jinqiu Sun, and Yanning Zhang.
\newblock You only need one color space: An efficient network for low-light image enhancement, 2024.

\bibitem{source18}
Yonghua Zhang, Jiawan Zhang, and Xiaojie Guo.
\newblock Kindling the darkness: A practical low-light image enhancer.
\newblock In {\em Proceedings of the 27th ACM International Conference on Multimedia}, MM '19, page 1632–1640, New York, NY, USA, 2019. Association for Computing Machinery.

\bibitem{source14}
Zhao Zhang, Huan Zheng, Richang Hong, Mingliang Xu, Shuicheng Yan, and Meng Wang.
\newblock Deep color consistent network for low-light image enhancement.
\newblock In {\em Proceedings of the IEEE/CVF Conference on Computer Vision and Pattern Recognition (CVPR)}, pages 1899--1908, June 2022.

\bibitem{source19}
Shen Zheng, Yiling Ma, Jinqian Pan, Changjie Lu, and Gaurav Gupta.
\newblock Low-light image and video enhancement: A comprehensive survey and beyond, 2022.

\end{thebibliography}

\appendix

\section{NSGA-II Based Multi-Objective Optimization }
\label{nsga}
\subsection{NSGA-II–Based Multi‐Objective Optimization}

Our core optimization engine is the NSGA-II (Non‐dominated Sorting Genetic Algorithm II) framework, which we adapt to jointly maximize image entropy and minimize deep‐feature distortion plus a brightness penalty. The main components are:

\begin{enumerate}
  \item \textbf{Representation.}  
    Each individual in the population encodes three real‐valued parameters 
    \(\mathbf{e} = [b, c, \gamma]\), corresponding to brightness shift, contrast scale, and gamma correction.  
    Bounds are enforced via clipping: 
    \[
       b \in [-10,60],\quad c,\gamma \in [1,2].
    \]

  \item \textbf{Initialization.}  
    We uniformly sample \(N\) individuals within these bounds to form the initial population \(P_0\).

  \item \textbf{Fitness Evaluation.}  
    For each individual \(\mathbf{e}\) , we apply the three‐step enhancement operator on the input image \(I_0\) to obtain \(I_{\text{enhanced}} = E(I_0;\mathbf{e})\).  
    We then compute the bi‐objective fitness  
    \begin{align*}
F(\mathbf{e}) &= \bigl(f_1,\,f_2\bigr), \\
f_1 &= \text{Entropy}(I_{\text{enhanced}}), \\
f_2 &= \|\Phi(I_0)-\Phi(I_{\text{enhanced}})\|_2 + \text{BrightnessPenalty}(I_{\text{enhanced}}).
\end{align*}

    Entropy is Shannon’s measure on the pixel histogram, \(\Phi(\cdot)\) extracts VGG16 features, and the penalty discourages mean‐brightness outside \([0.35,0.7]\).

  \item \textbf{Non‐Dominated Sorting \& Crowding Distance.}  
    NSGA-II partitions the combined parent + offspring population into Pareto fronts by non‐dominated sorting. Within each front, individuals are assigned a crowding distance score that measures local density in objective space.

  \item \textbf{Selection.}  
    We use binary tournament selection based on Pareto rank and crowding distance (preferring lower rank and, among equals, higher crowding distance) to choose parents for the next generation.

  \item \textbf{Variation.}  
    Parents undergo bounded blend crossover (\(\alpha=0.5\)) and Gaussian mutation (\(\sigma = 0.1\times\) parameter range) with an adaptive mutation probability \(r_{\text{mut}}\) that doubles when brightness diversity \(\sigma_{\text{b}}<5\), capped at 0.5.

  \item \textbf{Replacement.}  
    Offspring are evaluated and merged with the current population. The next generation \(P_{g+1}\) is formed by selecting the top \(N\) individuals again via non‐dominated sorting and crowding distance.

  \item \textbf{Memetic Local Search.}  
    To refine solutions, we apply an 8‐step hill‐climbing local search on the top 20\% of \(P_{g+1}\), accepting neighbor parameters that improve the primary objective or tie the primary while lowering the secondary.

  \item \textbf{Final Solution Selection.}  
    After \(G\) generations, NSGA-II yields a Pareto front of non‐dominated individuals. We select a single representative by ranking the front first by descending entropy \(f_1\), then by ascending secondary objective \(f_2\). In our DEAP implementation this corresponds to tools.selBest(pop,1)on the multi‐objective fitness.
\end{enumerate}

By integrating NSGA-II’s powerful Pareto‐based search with GPU-accelerated fitness evaluation and a memetic refinement phase, our algorithm efficiently discovers enhancement parameters that balance perceptual richness and semantic preservation on a per‐image basis—without any supervised training.
\newpage
\section{Summary of State-of-the-Art LLIE Methods}
\label{literature}

\begin{table}[htbp]
\centering
\caption{Summary of State-of-the-Art LLIE Methods}
\label{lit-summary}
\small
\renewcommand{\arraystretch}{1.1}
\setlength{\tabcolsep}{4pt}
\begin{tabularx}{\textwidth}{|>{\raggedright}p{2cm}|>{\raggedright}X|>{\raggedright}X|l|l|}
\hline
\textbf{Method} & \textbf{Key Feature} & \textbf{Strengths} & \textbf{Best On} & \textbf{Source} \\ 
\hline
Contrast-Driven Neural Network (CDNN) & 
Contrastive Encoder, Residual blocks, Multi-stage training & 
Color correction, over-enhancement mitigation & 
LOL + Underwater & 
\cite{source1} \\
\hline
MambaLLIE & 
Global-local Mamba architecture & 
Brightness preservation, detail restoration & 
LOL, MUSIQ, NIMA & 
\cite{source10} \\
\hline
Retinexformer & 
ORF framework, IG-MSA & 
Artifact resilience, Transformer-based generalization & 
13 datasets & 
\cite{source12} \\
\hline
CIDNet & 
Trainable HVI + LCA & 
Brightness-color decoupling, low FLOPs & 
11 datasets & 
\cite{source13} \\
\hline
Retinex-RNet & 
Residual attention, Illumination modeling & 
Fine-grained local enhancement & 
LOL, LIME, DICM & 
\cite{source3} \\
\hline
PairLIE & 
Paired instance learning & 
Accurate reflectance decomposition & 
LOL, SICE & 
\cite{source7} \\
\hline
PAS-MEF & 
PCA + Saliency fusion & 
Balanced exposure, HVS-like behavior & 
MEF scenarios & 
\cite{source11} \\
\hline
Fast Fusion LLIE & 
Histogram + Bilateral fusion & 
Local detail preservation, fast processing & 
NIQE, SSEQ & 
\cite{source6} \\
\hline
Contrast RGB & 
Channel-wise enhancement & 
High SSIM on dark images & 
ExDark & 
\cite{source2} \\
\hline
HSV Method & 
HSV Power-law + PCA & 
Color balance, minimal distortion & 
AG, UQI & 
\cite{source8} \\
\hline
\end{tabularx}
\end{table}


\section{Visual examples for unpaired image enhancement on unpaired dataset }
\label{visualizeunpaired}
\begin{figure}[H]
\centering
\setlength{\tabcolsep}{3pt} 
\renewcommand{\arraystretch}{1} 

\begin{tabular}{cc}
    \begin{subfigure}[t]{0.23\textwidth}
        \includegraphics[width=\linewidth]{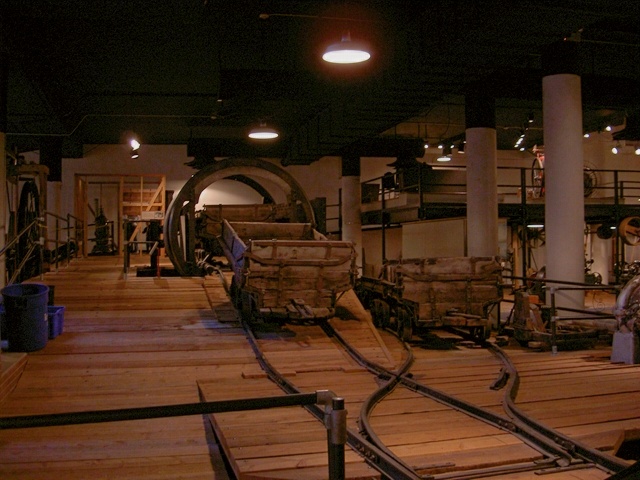}
        \caption*{Original}
    \end{subfigure} &
    \begin{subfigure}[t]{0.23\textwidth}
        \includegraphics[width=\linewidth]{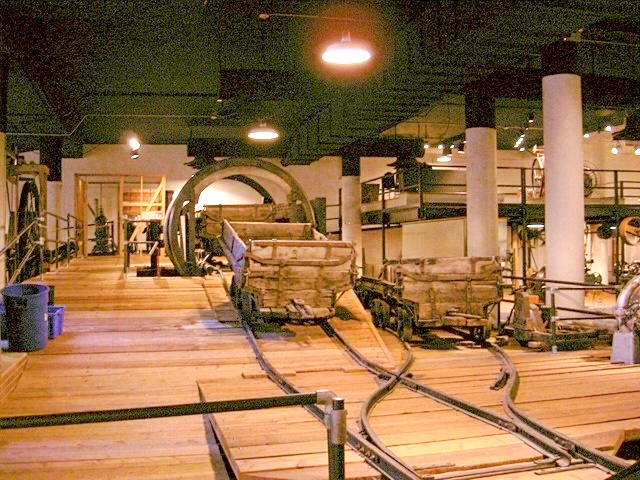}
        \caption*{Full Model}
    \end{subfigure} \\

    \begin{subfigure}[t]{0.23\textwidth}
        \includegraphics[width=\linewidth]{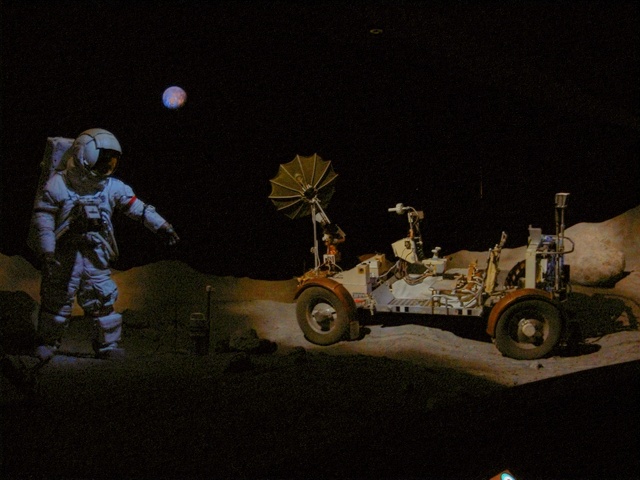}
        \caption*{Original}
    \end{subfigure} &
    \begin{subfigure}[t]{0.23\textwidth}
        \includegraphics[width=\linewidth]{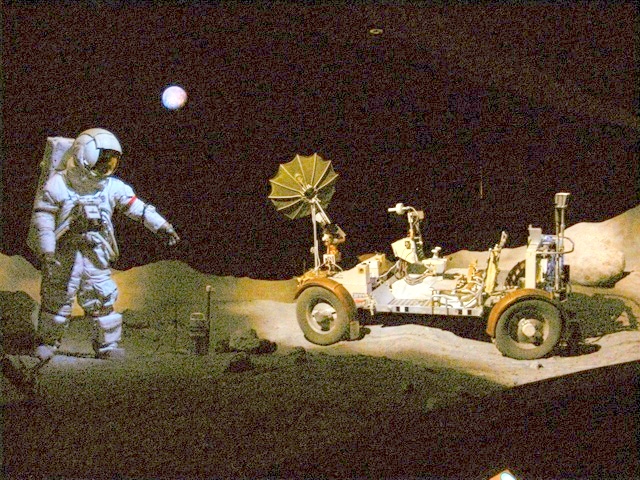}
        \caption*{Full Model}
    \end{subfigure} \\

    \begin{subfigure}[t]{0.23\textwidth}
        \includegraphics[width=\linewidth]{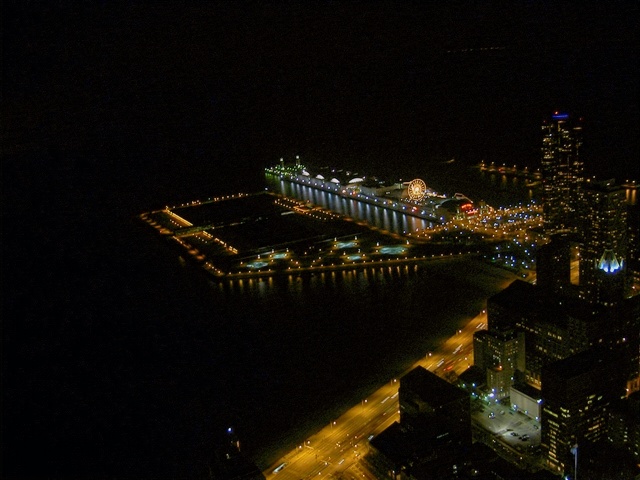}
        \caption*{Original}
    \end{subfigure} &
    \begin{subfigure}[t]{0.23\textwidth}
        \includegraphics[width=\linewidth]{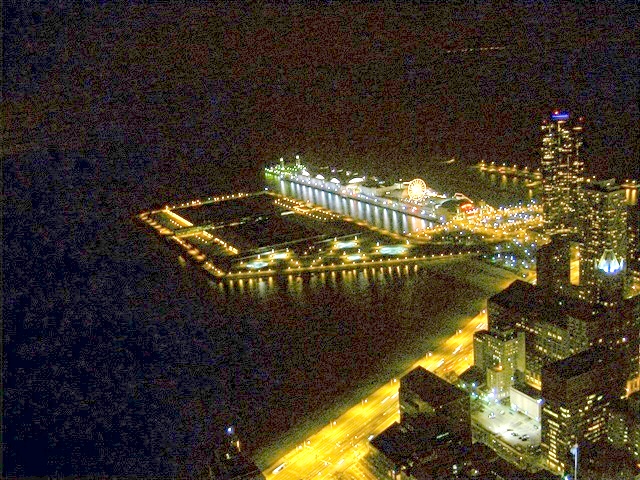}
        \caption*{Full Model}
    \end{subfigure} \\

    \begin{subfigure}[t]{0.23\textwidth}
        \includegraphics[width=\linewidth]{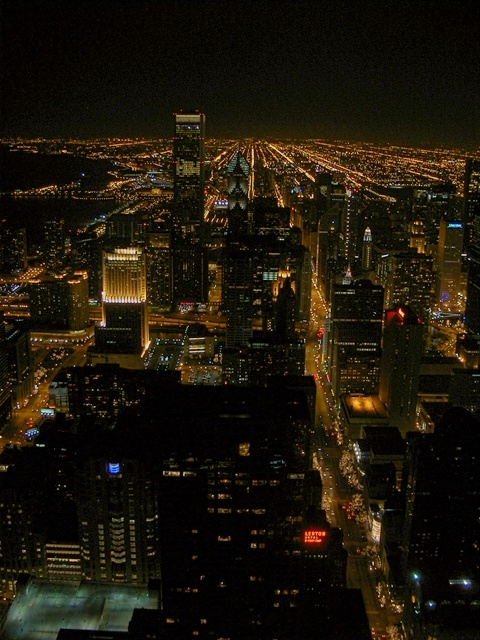}
        \caption*{Original}
    \end{subfigure} &
    \begin{subfigure}[t]{0.23\textwidth}
        \includegraphics[width=\linewidth]{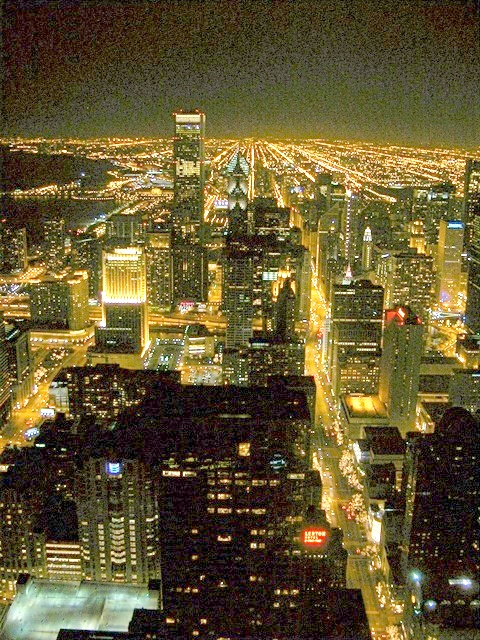}
        \caption*{Full Model}
    \end{subfigure}
\end{tabular}

\vspace{0.5em}
\caption{Example enhancement on the DICM dataset using our proposed method}
\label{fig:examples_dicm}
\end{figure}
\newpage
\begin{figure}[H]
\centering
\setlength{\tabcolsep}{3pt} 
\renewcommand{\arraystretch}{1} 

\begin{tabular}{cc}
    \begin{subfigure}[t]{0.23\textwidth}
        \includegraphics[width=\linewidth]{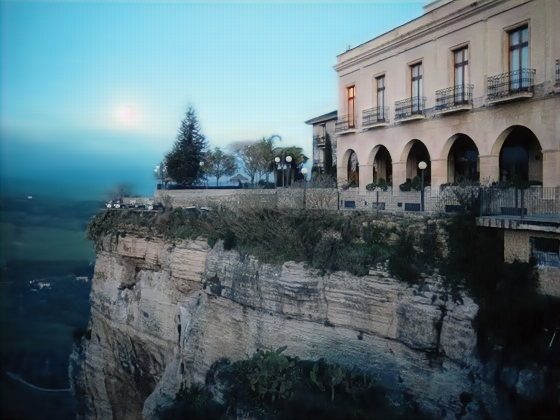}
        \caption*{Original}
    \end{subfigure} &
    \begin{subfigure}[t]{0.23\textwidth}
        \includegraphics[width=\linewidth]{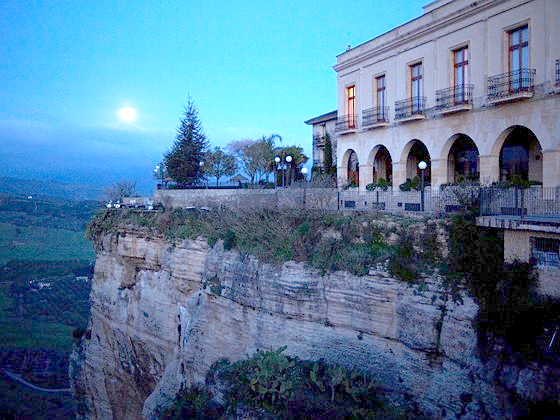}
        \caption*{Full Model}
    \end{subfigure} \\

    \begin{subfigure}[t]{0.23\textwidth}
        \includegraphics[width=\linewidth]{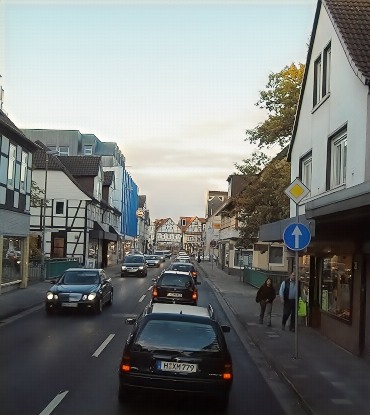}
        \caption*{Original}
    \end{subfigure} &
    \begin{subfigure}[t]{0.23\textwidth}
        \includegraphics[width=\linewidth]{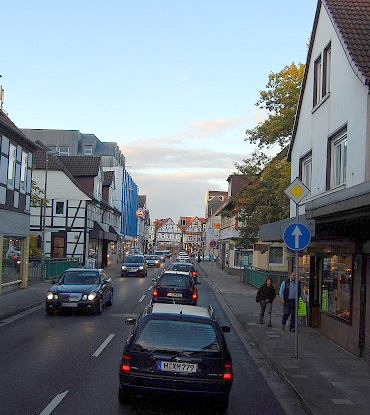}
        \caption*{Full Model}
    \end{subfigure} \\

    \begin{subfigure}[t]{0.23\textwidth}
        \includegraphics[width=\linewidth]{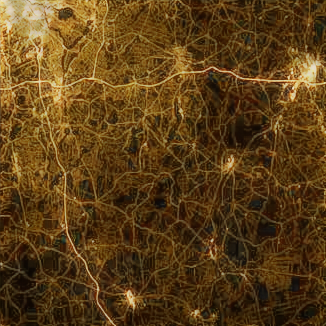}
        \caption*{Original}
    \end{subfigure} &
    \begin{subfigure}[t]{0.23\textwidth}
        \includegraphics[width=\linewidth]{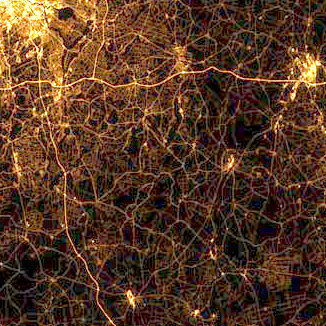}
        \caption*{Full Model}
    \end{subfigure} \\

    \begin{subfigure}[t]{0.23\textwidth}
        \includegraphics[width=\linewidth]{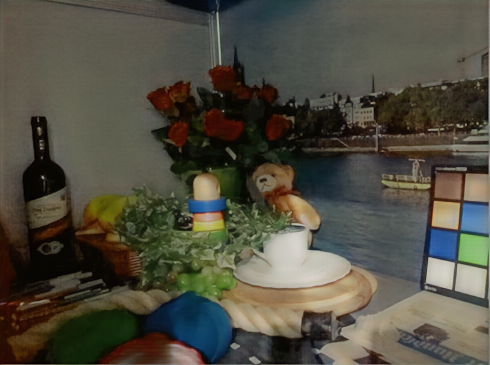}
        \caption*{Original}
    \end{subfigure} &
    \begin{subfigure}[t]{0.23\textwidth}
        \includegraphics[width=\linewidth]{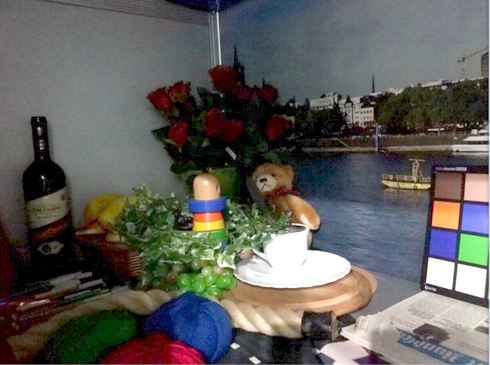}
        \caption*{Full Model}
    \end{subfigure}
\end{tabular}

\vspace{0.5em}
\caption{Example enhancement on the LIME dataset using our proposed method}
\label{fig:examples_lime}
\end{figure}
\newpage
\begin{figure}[H]
\centering
\setlength{\tabcolsep}{3pt} 
\renewcommand{\arraystretch}{1} 

\begin{tabular}{cc}
    \begin{subfigure}[t]{0.23\textwidth}
        \includegraphics[width=\linewidth]{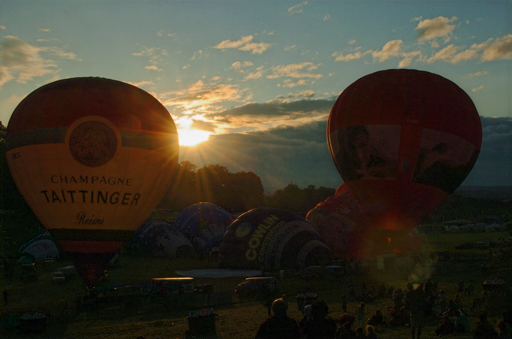}
        \caption*{Original}
    \end{subfigure} &
    \begin{subfigure}[t]{0.23\textwidth}
        \includegraphics[width=\linewidth]{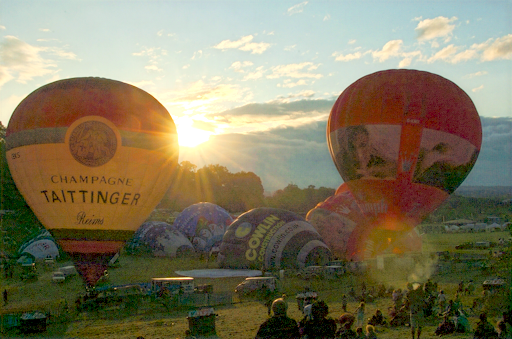}
        \caption*{Full Model}
    \end{subfigure} \\

    \begin{subfigure}[t]{0.23\textwidth}
        \includegraphics[width=\linewidth]{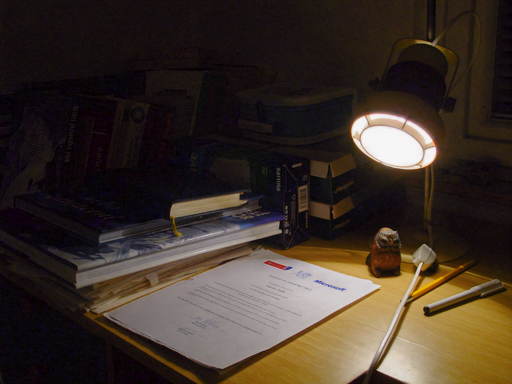}
        \caption*{Original}
    \end{subfigure} &
    \begin{subfigure}[t]{0.23\textwidth}
        \includegraphics[width=\linewidth]{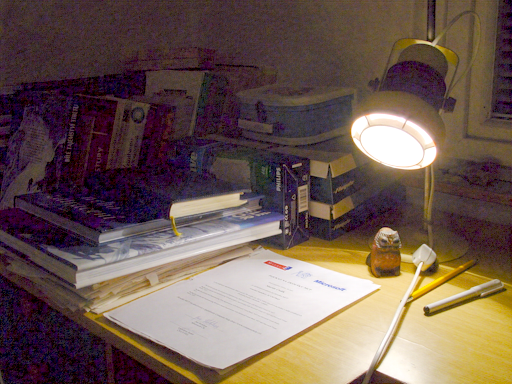}
        \caption*{Full Model}
    \end{subfigure} \\

    \begin{subfigure}[t]{0.23\textwidth}
        \includegraphics[width=\linewidth]{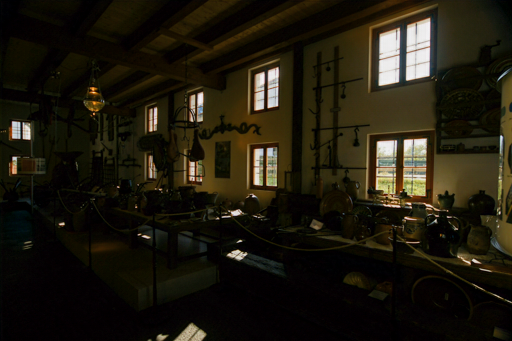}
        \caption*{Original}
    \end{subfigure} &
    \begin{subfigure}[t]{0.23\textwidth}
        \includegraphics[width=\linewidth]{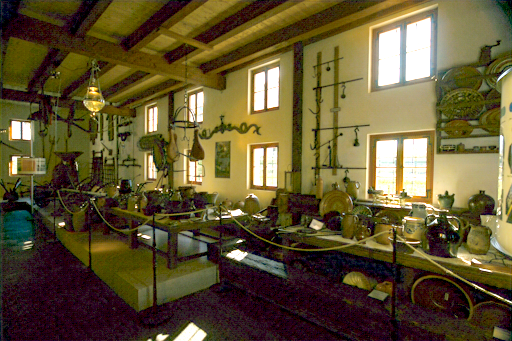}
        \caption*{Full Model}
    \end{subfigure} \\

    \begin{subfigure}[t]{0.23\textwidth}
        \includegraphics[width=\linewidth]{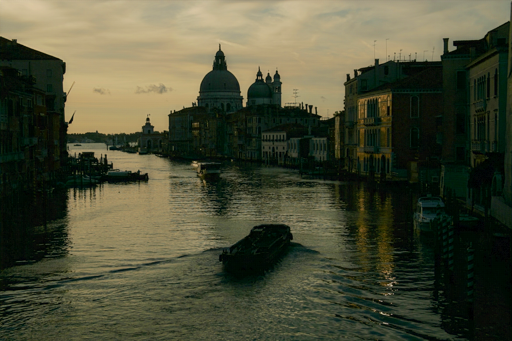}
        \caption*{Original}
    \end{subfigure} &
    \begin{subfigure}[t]{0.23\textwidth}
        \includegraphics[width=\linewidth]{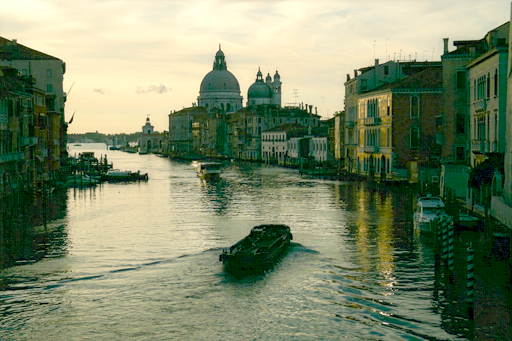}
        \caption*{Full Model}
    \end{subfigure}
\end{tabular}

\vspace{0.5em}
\caption{Example enhancement on the MEF dataset using our proposed method}
\label{fig:examples_MEF}
\end{figure}
\begin{figure}[H]
\centering
\setlength{\tabcolsep}{3pt} 
\renewcommand{\arraystretch}{1} 

\begin{tabular}{cc}
    \begin{subfigure}[t]{0.23\textwidth}
        \includegraphics[width=\linewidth]{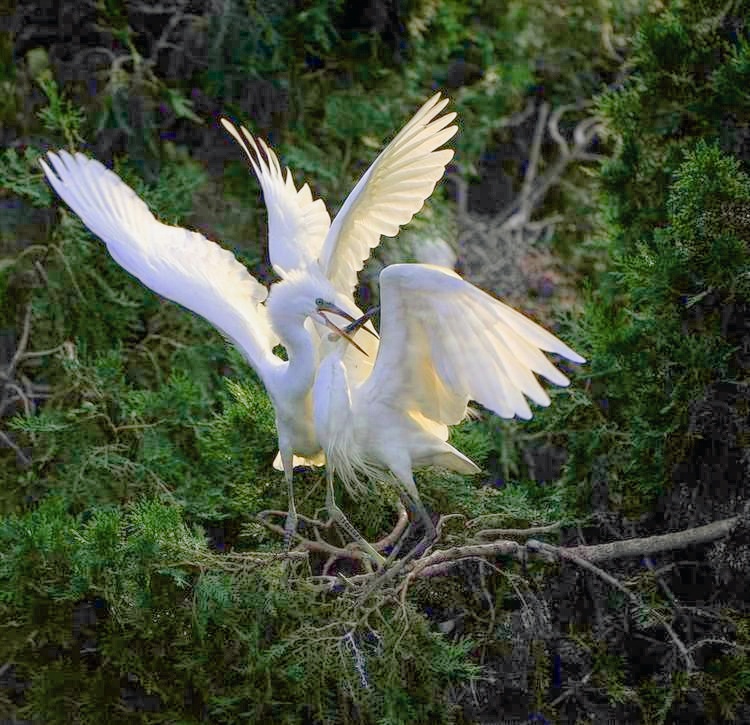}
        \caption*{Original}
    \end{subfigure} &
    \begin{subfigure}[t]{0.23\textwidth}
        \includegraphics[width=\linewidth]{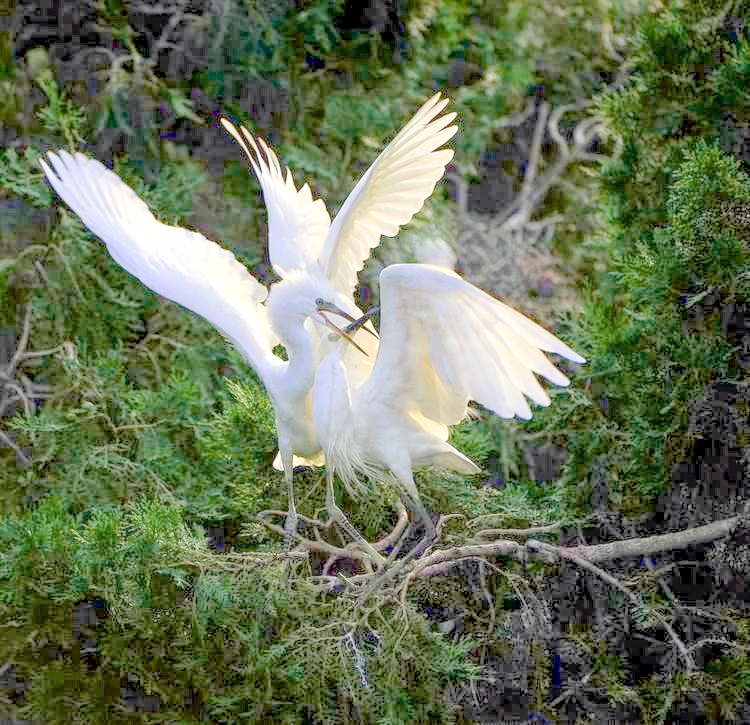}
        \caption*{Full Model}
    \end{subfigure} \\

    \begin{subfigure}[t]{0.23\textwidth}
        \includegraphics[width=\linewidth]{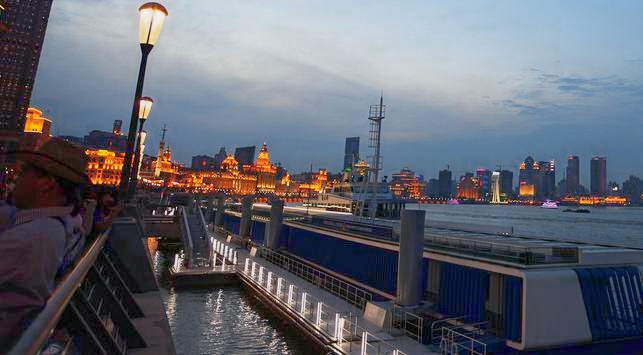}
        \caption*{Original}
    \end{subfigure} &
    \begin{subfigure}[t]{0.23\textwidth}
        \includegraphics[width=\linewidth]{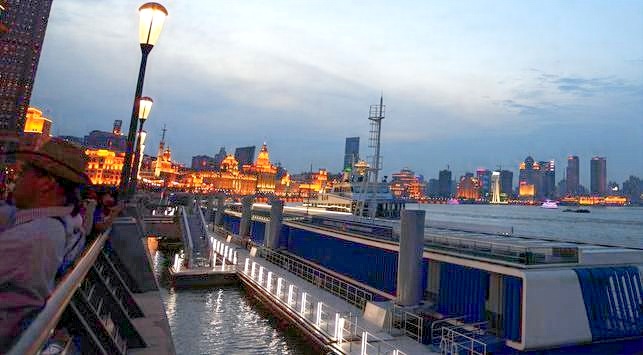}
        \caption*{Full Model}
    \end{subfigure} \\

    \begin{subfigure}[t]{0.23\textwidth}
        \includegraphics[width=\linewidth]{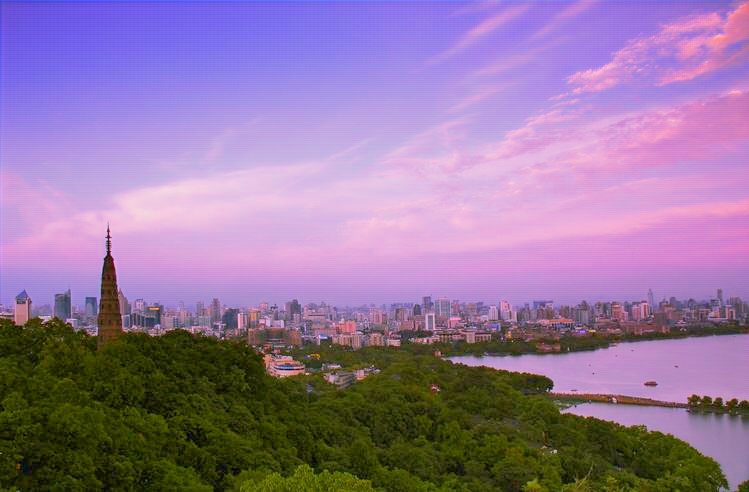}
        \caption*{Original}
    \end{subfigure} &
    \begin{subfigure}[t]{0.23\textwidth}
        \includegraphics[width=\linewidth]{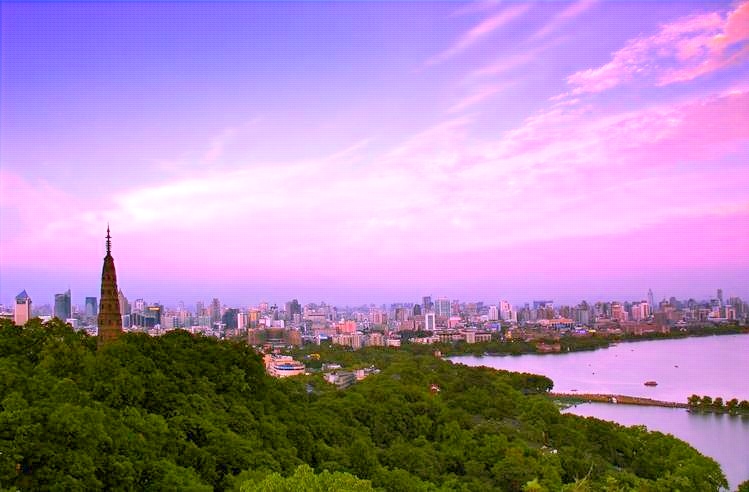}
        \caption*{Full Model}
    \end{subfigure} \\

    \begin{subfigure}[t]{0.23\textwidth}
        \includegraphics[width=\linewidth]{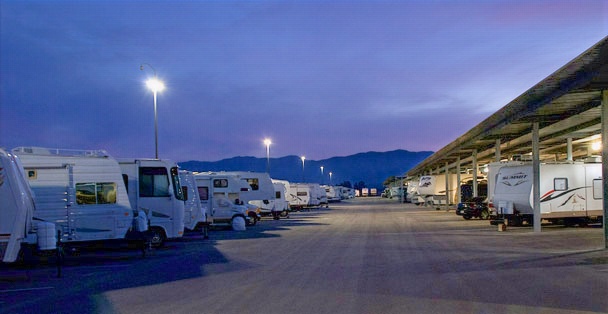}
        \caption*{Original}
    \end{subfigure} &
    \begin{subfigure}[t]{0.23\textwidth}
        \includegraphics[width=\linewidth]{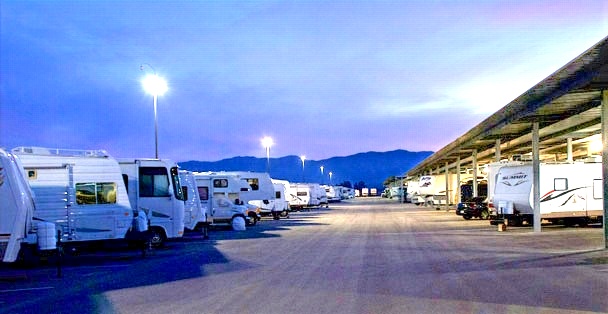}
        \caption*{Full Model}
    \end{subfigure}
\end{tabular}

\vspace{0.5em}
\caption{Example enhancement on the NPE dataset using our proposed method}
\label{fig:examples_NPE}
\end{figure}
\newpage
\begin{figure}[H]
\centering
\setlength{\tabcolsep}{3pt} 
\renewcommand{\arraystretch}{1} 

\begin{tabular}{cc}
    \begin{subfigure}[t]{0.23\textwidth}
        \includegraphics[width=\linewidth]{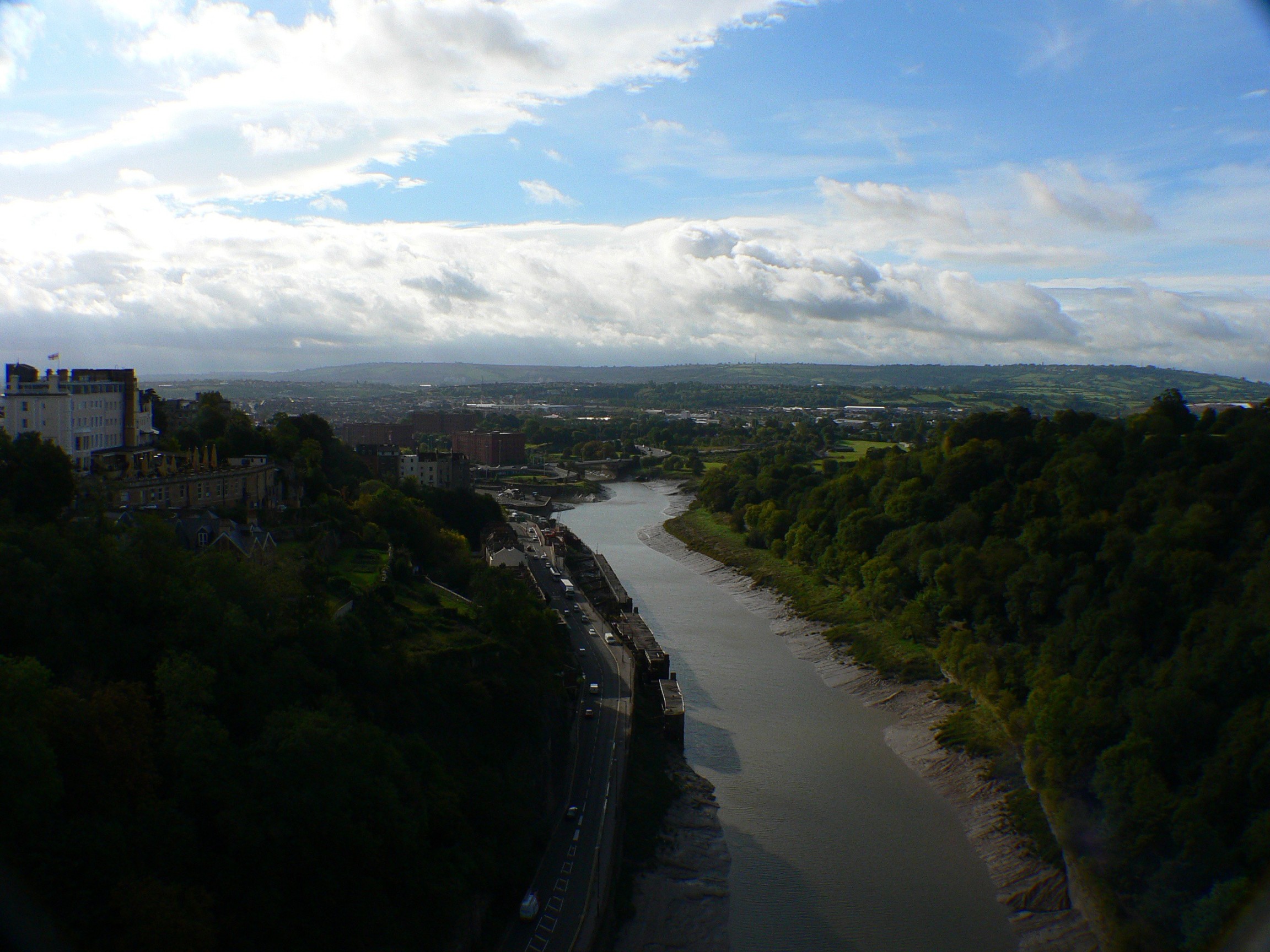}
        \caption*{Original}
    \end{subfigure} &
    \begin{subfigure}[t]{0.23\textwidth}
        \includegraphics[width=\linewidth]{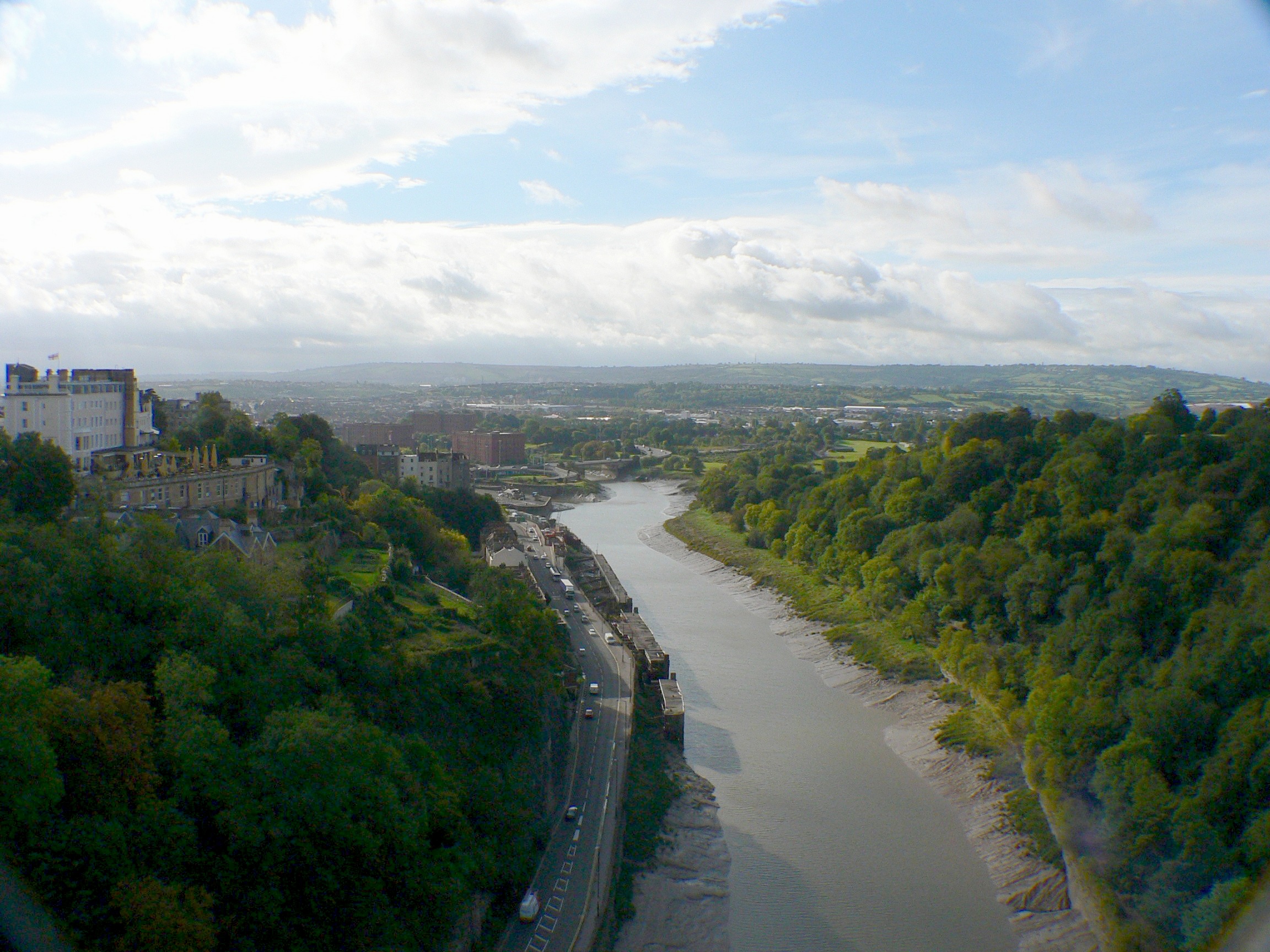}
        \caption*{Full Model}
    \end{subfigure} \\

    \begin{subfigure}[t]{0.23\textwidth}
        \includegraphics[width=\linewidth]{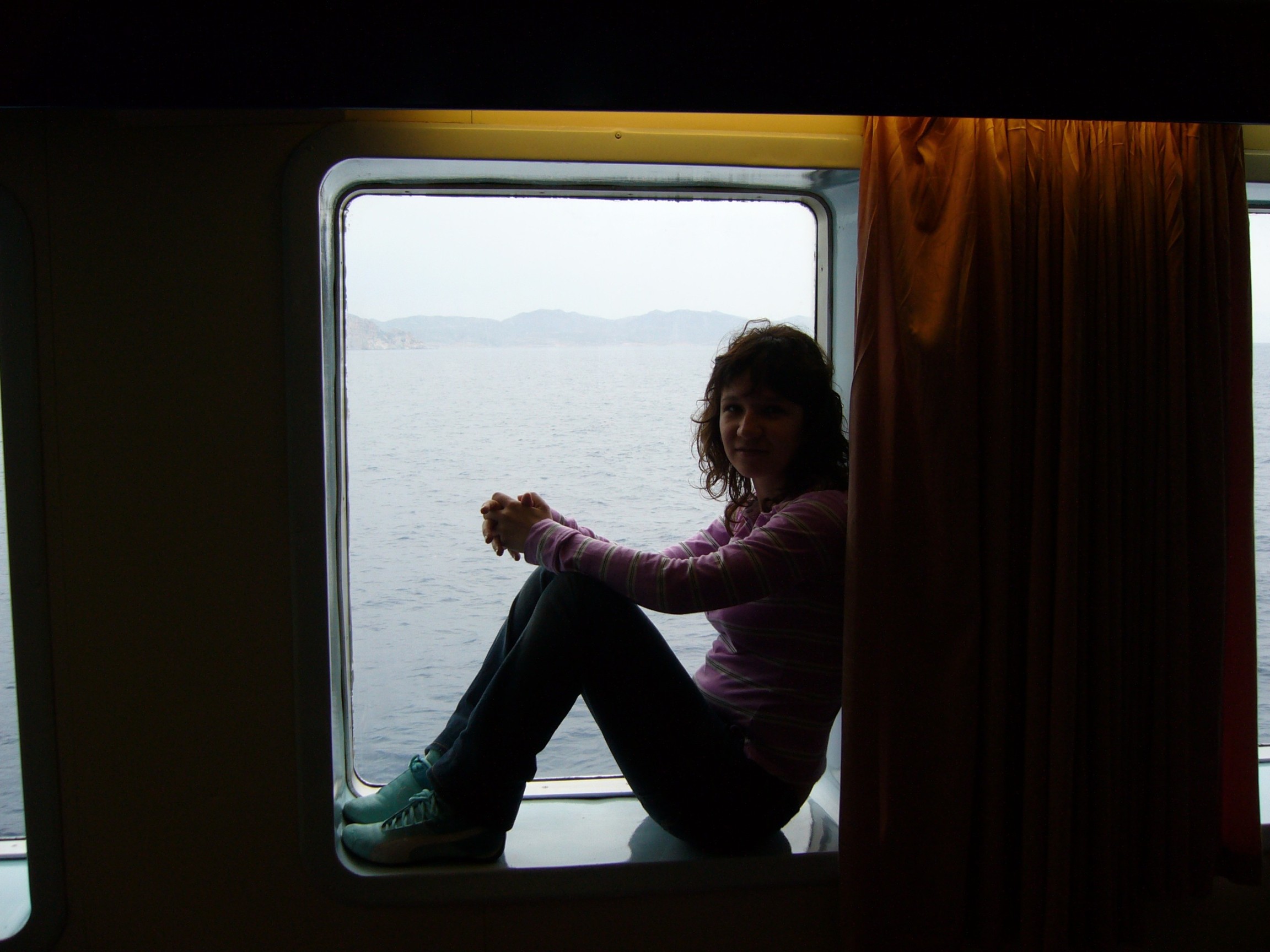}
        \caption*{Original}
    \end{subfigure} &
    \begin{subfigure}[t]{0.23\textwidth}
        \includegraphics[width=\linewidth]{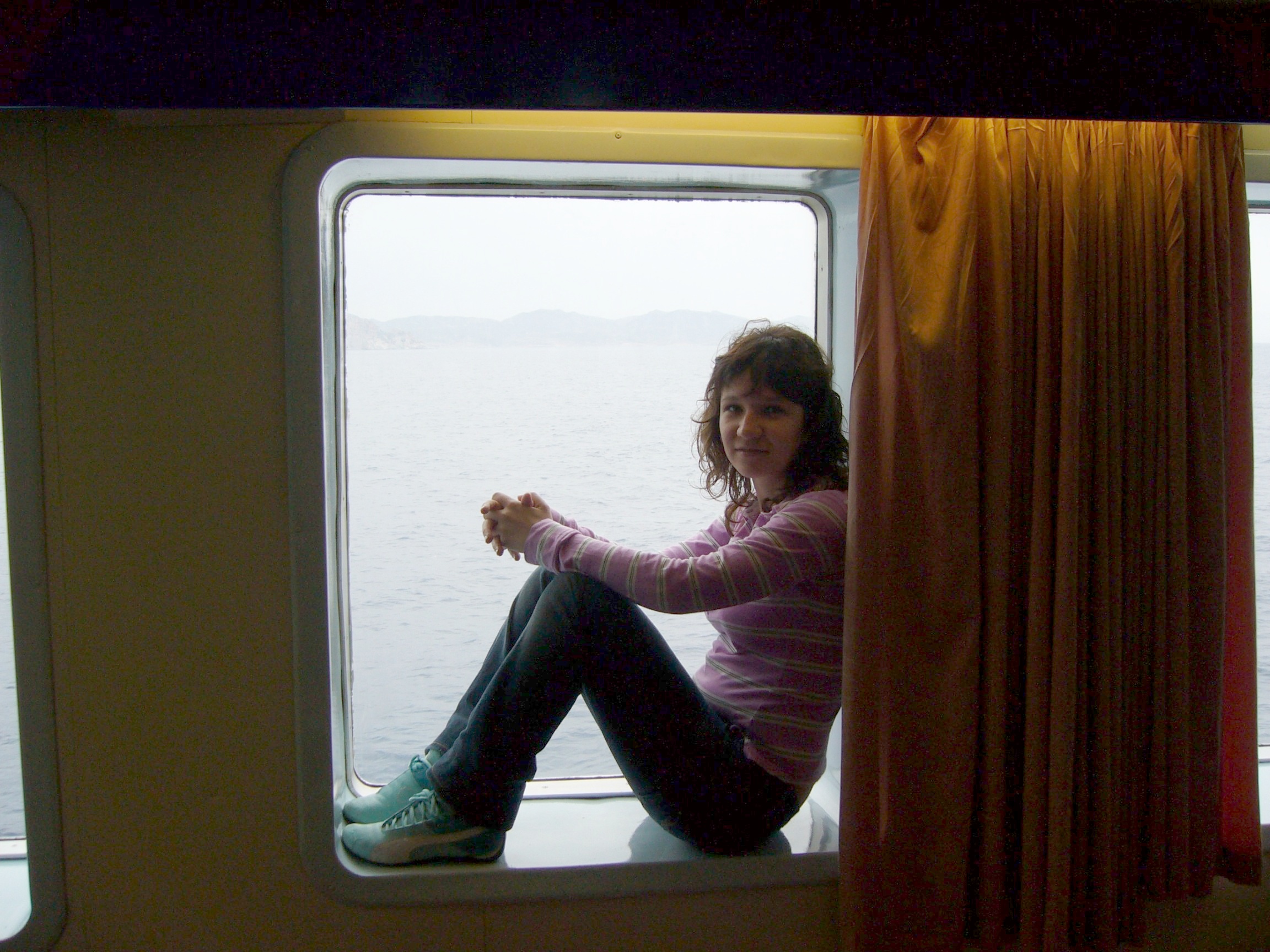}
        \caption*{Full Model}
    \end{subfigure} \\

    \begin{subfigure}[t]{0.23\textwidth}
        \includegraphics[width=\linewidth]{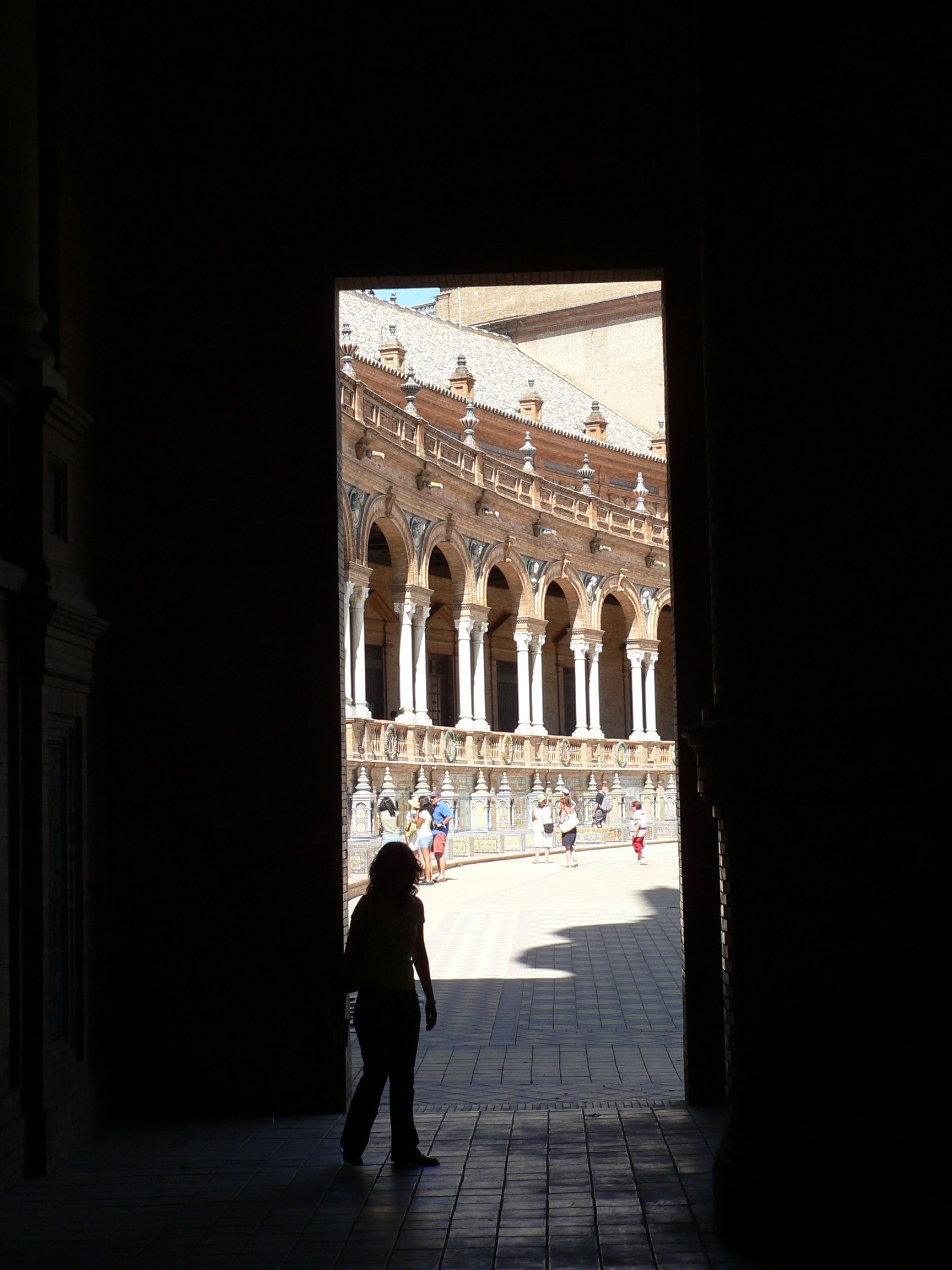}
        \caption*{Original}
    \end{subfigure} &
    \begin{subfigure}[t]{0.23\textwidth}
        \includegraphics[width=\linewidth]{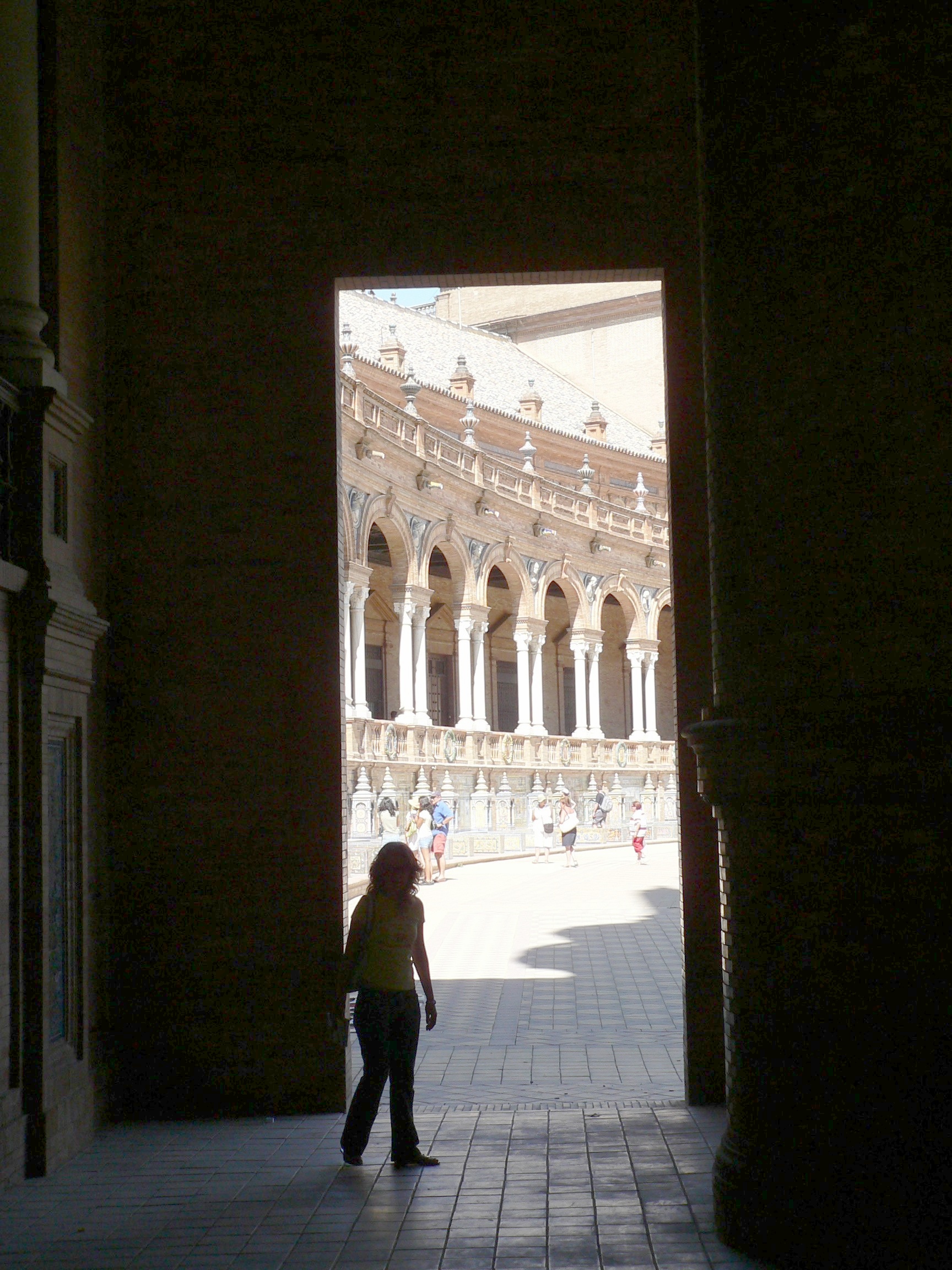}
        \caption*{Full Model}
    \end{subfigure} \\

    \begin{subfigure}[t]{0.23\textwidth}
        \includegraphics[width=\linewidth]{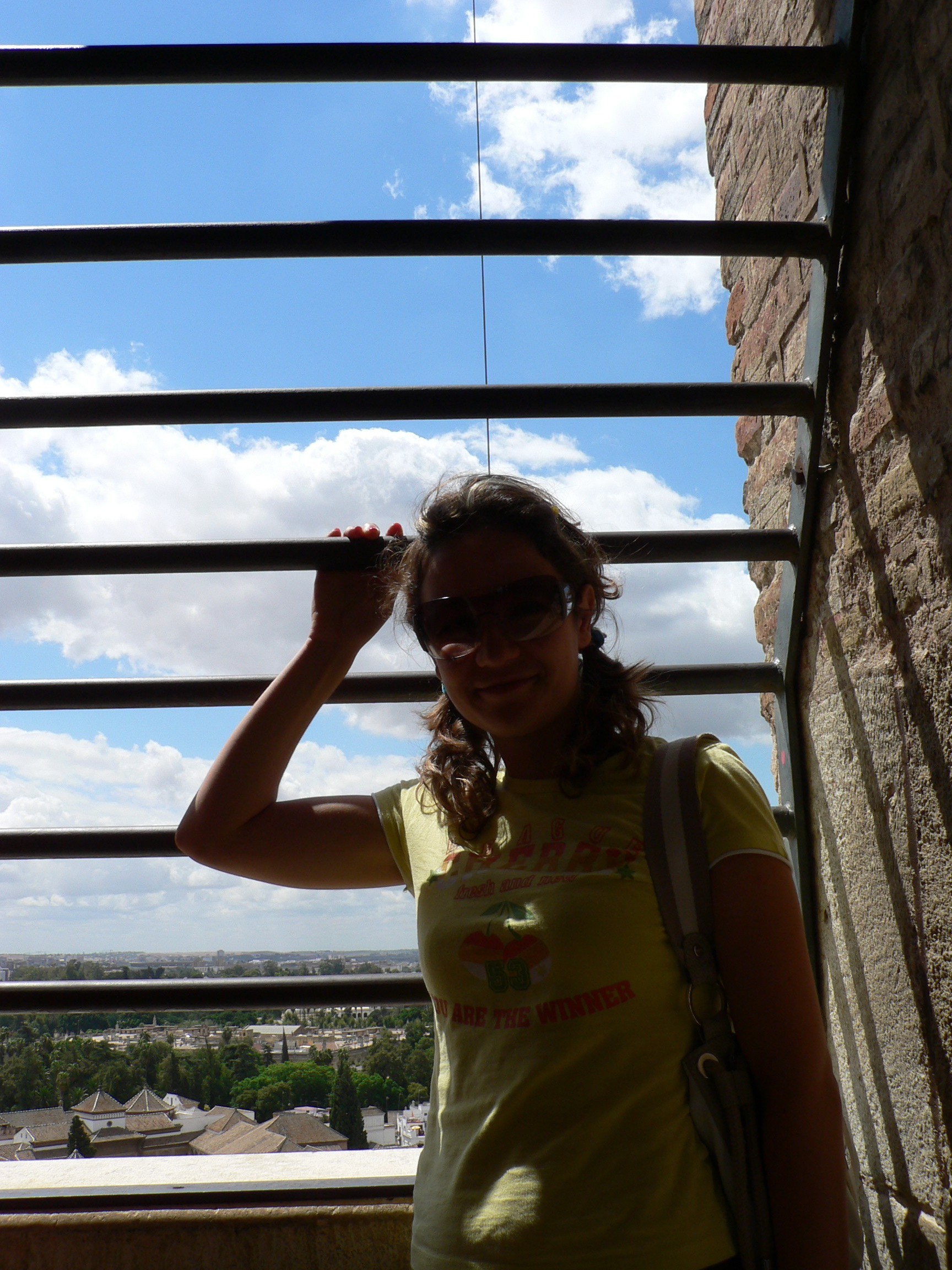}
        \caption*{Original}
    \end{subfigure} &
    \begin{subfigure}[t]{0.23\textwidth}
        \includegraphics[width=\linewidth]{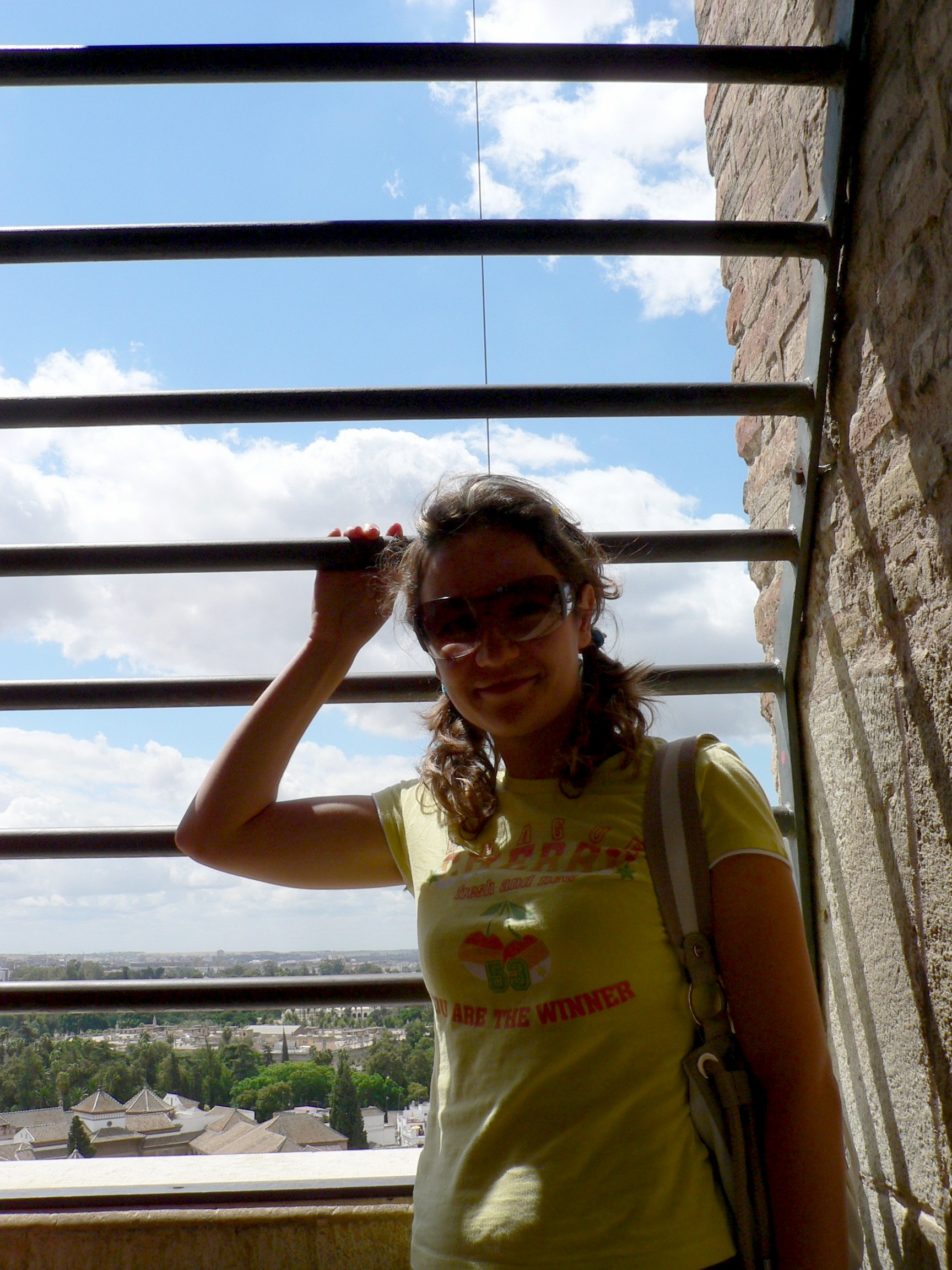}
        \caption*{Full Model}
    \end{subfigure}
\end{tabular}

\vspace{0.5em}
\caption{Example enhancement on the vv dataset. using our proposed method}
\label{fig:examples_vv}
\end{figure}

\end{document}